\documentclass[10pt,journal,compsoc]{IEEEtran}

\ifCLASSOPTIONcompsoc
  \usepackage[nocompress]{cite}
\else
  \usepackage{cite}
\fi

\ifCLASSINFOpdf
\else
\fi

\usepackage{url}
\usepackage[table,dvipsnames]{xcolor}
\usepackage{booktabs}
\usepackage{multicol}
\usepackage{multirow}
\usepackage{color}
\usepackage{caption}
\usepackage{subcaption}
\usepackage{hhline}
\usepackage{pifont}
\usepackage{threeparttable}
\usepackage{makecell}
\usepackage{algorithm} 
\usepackage{listings}
\usepackage{amsfonts,amssymb}
\usepackage{amsmath}
\usepackage{graphicx}
\usepackage{lipsum}
\usepackage{arydshln}
\usepackage[accsupp]{axessibility}
\usepackage{enumitem}
\usepackage{float}

\def\ie{\emph{i.e.}}
\def\eg{\emph{e.g.}}

\def\etal{{\em et al.~}}
\def\vs{{\em vs.}}

\newcommand{\tabincell}[2]{\begin{tabular}{@{}#1@{}}#2\end{tabular}}

\newcommand{\myPara}[1]{\vspace{.05in}\noindent\textbf{#1.}}
\newlength\savedwidth
\newcommand\whline{\noalign{\global\savedwidth\arrayrulewidth\global\arrayrulewidth 0.8pt}\hline\noalign{\global\arrayrulewidth\savedwidth}}
\usepackage{verbatim}
\newcommand{\cmark}{\ding{51}}

\newcommand{\xmarkg}{\textcolor{lightgray}{\ding{55}}}
\newcommand{\gr}{\rowcolor[gray]{.95}}
\definecolor{mylb}{RGB}{229, 247, 255}
\definecolor{mygray}{gray}{.95}
\definecolor{darkred}{rgb}{0.55, 0.0, 0.0}
\definecolor{myred}{HTML}{ED058C}

\newcolumntype{a}{>{\columncolor{mylb}}c}
\usepackage{amsthm}
\makeatletter
\DeclareRobustCommand{\iscircle}{\mathord{\mathpalette\is@circle\relax}}
\newcommand\is@circle[2]{%
  \begingroup
  \sbox\z@{\raisebox{\depth}{$\m@th#1\bigcirc$}}%
  \sbox\tw@{$#1\square$}%
  \resizebox{!}{\ht\tw@}{\usebox{\z@}}%
  \endgroup
}

\newcommand{\tablestyle}[2]{\setlength{\tabcolsep}{#1}\renewcommand{\arraystretch}{#2}\centering\footnotesize}

\makeatother

\begin{document}
\title{MetaFormer Baselines for Vision}

\author{Weihao Yu,
        Chenyang Si,
        Pan Zhou,
        Mi Luo,
        Yichen Zhou,
        Jiashi Feng,\\
        Shuicheng Yan,~\IEEEmembership{Fellow,~IEEE,}
        and Xinchao Wang,~\IEEEmembership{Senior Member,~IEEE}%
\IEEEcompsocitemizethanks{
\IEEEcompsocthanksitem This work was partially performed when Weihao Yu was a research intern at Sea AI Lab.
\IEEEcompsocthanksitem Weihao Yu, Mi Luo and Xinchao Wang are with National University of Singapore. \protect\\
Emails: weihaoyu@u.nus.edu, xinchao@nus.edu.sg.
\IEEEcompsocthanksitem Chenyang Si, Pan Zhou, Yichen Zhou, Jiashi Feng, and Shuicheng Yan are with Sea AI Lab. Email: yansc@sea.com.
\IEEEcompsocthanksitem Corresponding authors: Xinchao Wang and Shuicheng Yan.
}%

}

\markboth{IEEE Transactions on Pattern Analysis and Machine Intelligence}%
{Yu \etal{}: MetaFormer Baselines for Vision}

\IEEEtitleabstractindextext{%
\begin{abstract}
MetaFormer, the abstracted architecture of Transformer, has been found to play a significant role in achieving competitive performance. In this paper, we further explore the capacity of MetaFormer, again, by migrating our focus away from the token mixer design: we introduce several baseline models under MetaFormer using the most basic or common mixers, and demonstrate their gratifying performance. We summarize our observations as follows:
\begin{enumerate}[label=(\arabic*)]
\item \textbf{MetaFormer ensures solid lower bound of performance}. By merely adopting identity mapping as the token mixer, the MetaFormer model, termed \textit{IdentityFormer}, achieves $>$80\% accuracy on ImageNet-1K.
\item \textbf{MetaFormer works well with arbitrary token mixers.} When specifying the token mixer as even a random matrix to mix tokens, the resulting model \textit{RandFormer} yields an accuracy of $>$81\%, outperforming IdentityFormer. Rest assured of MetaFormer's results when new token mixers are adopted.
\item \textbf{MetaFormer effortlessly offers state-of-the-art results.}  With just conventional token mixers dated back five years ago, the models instantiated from MetaFormer already beat state of the art.
\begin{enumerate}[label=(\alph*)]
\item \textbf{ConvFormer outperforms ConvNeXt}. Taking the common depthwise separable convolutions as the token mixer, the model termed \textit{ConvFormer}, which can be regarded as pure CNNs, outperforms the strong CNN model ConvNeXt.
\item \textbf{CAFormer sets new record on ImageNet-1K}. By simply applying depthwise separable convolutions as token mixer in the bottom stages and vanilla self-attention in the top stages, the resulting model \textit{CAFormer} sets a new record on ImageNet-1K: it achieves an accuracy of 85.5\% at $224 \times 224$ resolution, under normal supervised training without external data or distillation.
\end{enumerate}
\end{enumerate}
In our expedition to probe MetaFormer, we also find that a new activation, \textit{StarReLU}, reduces 71\% FLOPs of activation compared with commonly-used GELU yet achieves better performance. Specifically, StarReLU is a variant of Squared ReLU dedicated to alleviating distribution shift. We expect StarReLU to find great potential in MetaFormer-like models alongside other neural networks. Code and models are available at \url{https://github.com/sail-sg/metaformer}.
\end{abstract}

}

\maketitle

\IEEEdisplaynontitleabstractindextext

\IEEEpeerreviewmaketitle
\begin{figure*}[ht]
\hsize=\textwidth
\centering
\begin{subfigure}[b]{0.39\textwidth}
    \centering
    \includegraphics[width=1\textwidth]{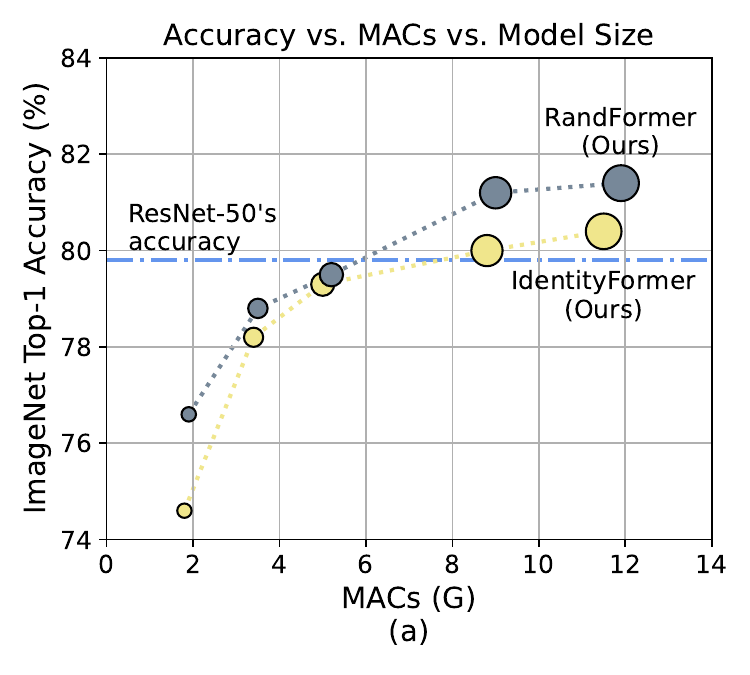}
\end{subfigure}    
\hspace{-0.2in}
\begin{subfigure}[b]{0.39\textwidth}
     \centering
     \includegraphics[width=1\textwidth]{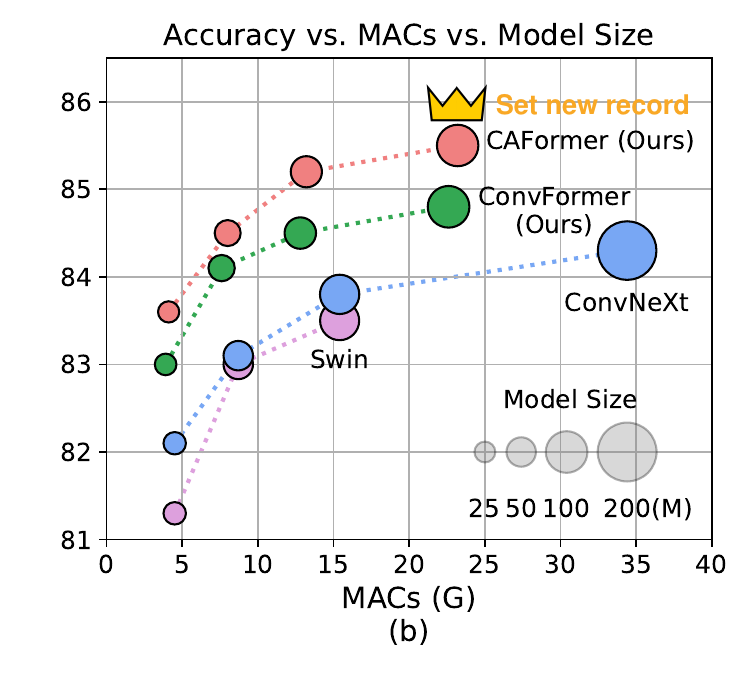}
\end{subfigure}
\caption{
\textbf{Performance of MetaFormer baselines and other state-of-the-art models on ImageNet-1K at $224^2$ resolution.} The architectures of our proposed models are shown in Figure \ref{fig:overall_framework}. (a) IdentityFormer/RandFormer achieve over 80\%/81\% accuracy, indicating MetaFormer has solid lower bound of performance and works well on arbitrary token mixers. The accuracy of well-trained ResNet-50 \cite{resnet} is from \cite{rsb}. (b) Without novel token mixers, pure CNN-based ConvFormer outperforms ConvNeXt \cite{convnext},
while CAFormer sets a new record of 85.5\% accuracy on ImageNet-1K at $224^2$ resolution under normal supervised training without external data or distillation.
}
\label{fig:first_figure}
\end{figure*}

\begin{figure*}[t]
  \centering
   \includegraphics[width=1\linewidth]{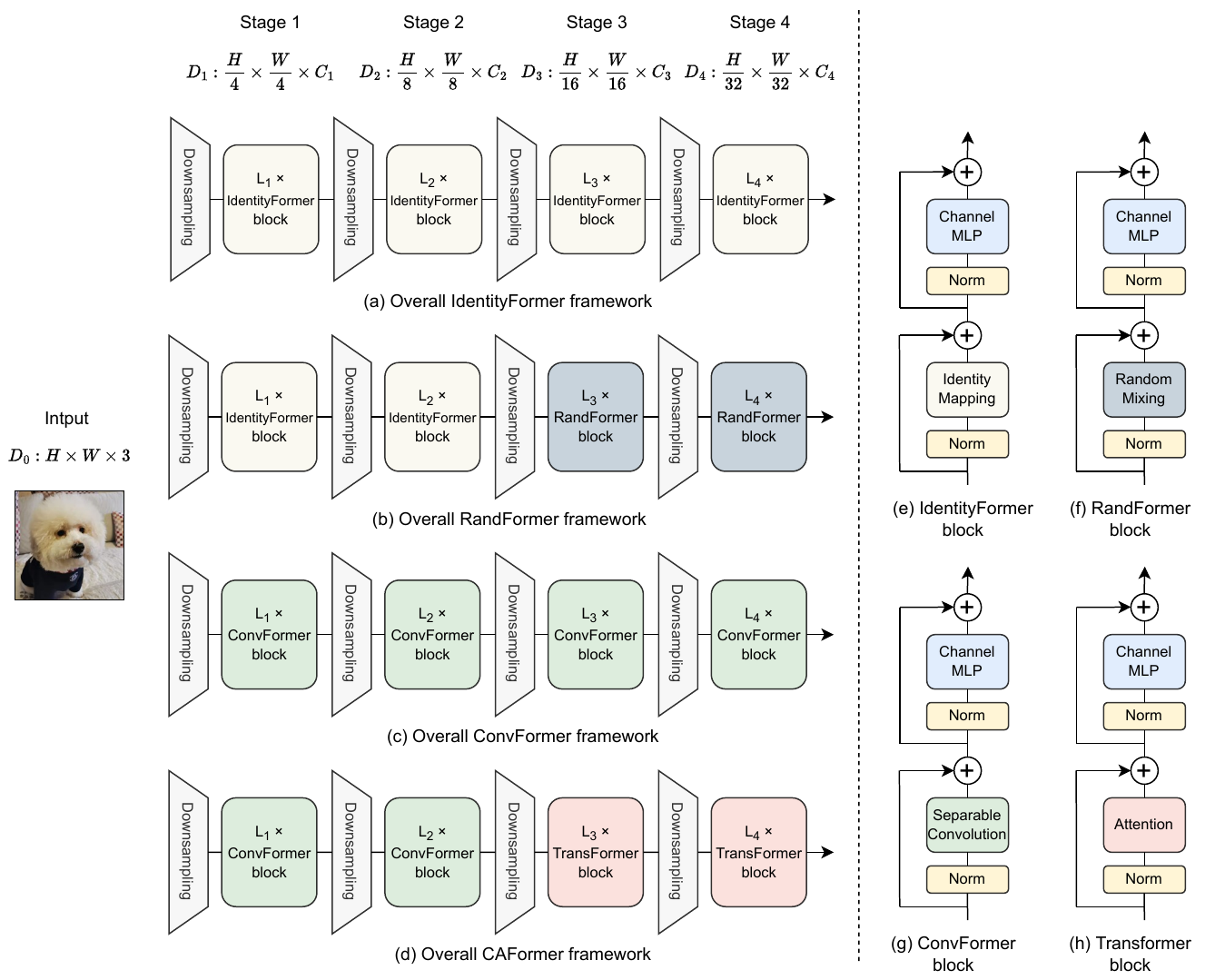}
   \caption{\textbf{(a-d)  Overall frameworks of IdentityFormer, RandFormer, ConvFormer and CAFormer.} Similar to \cite{resnet, pvt, swin}, the models adopt hierarchical architecture of 4 stages, and stage $i$ has  $L_i$ blocks with feature dimension $D_i$. Each downsampling module is implemented by a layer of convolution. The first downsampling has kernel size of 7 and stride of 4, while the last three ones have kernel size of 3 and stride of 2. \textbf{(e-h) Architectures of IdentityFormer, RandFormer, ConvFormer and Transformer blocks}, which have token mixer of identity mapping, global random mixing (Equation \ref{eqn:random_mixing}), separable depthwise convolutions \cite{chollet2017xception, mamalet2012simplifying, mobilenetv2} (Equation \ref{eqn:convolutions}) or vanilla self-attention \cite{transformer},  respectively. }
   \label{fig:overall_framework}
\end{figure*}

\IEEEraisesectionheading{\section{Introduction}\label{sec:introduction}}
\IEEEPARstart{I}{n} recent years, Transformers \cite{transformer} have demonstrated unprecedented success in various computer vision tasks \cite{igpt, vit, detr, setr}. The competence of Transformers has been long attributed to its attention module. As such, many attention-based token mixers \cite{t2t, tnt, pvt, swin, refiner} have been proposed in the aim to strengthen the Vision Transformers~(ViTs)~\cite{vit}. Nevertheless, some work~\cite{mlp-mixer, lee2021fnet, zhao2021battle, han2021connection, rao2021global} found that, by replacing the attention module in Transformers with simple operators like spatial MLP \cite{mlp-mixer, resmlp, tay2021synthesizer} or Fourier transform \cite{lee2021fnet}, the resultant models still produce encouraging performance.

Along this line, the work \cite{metaformer} abstracts Transformer into a general architecture termed \textit{MetaFormer}, and hypothesizes that it is MetaFormer that plays an essential role for models in achieving competitive performance. To verify this hypothesis, \cite{metaformer} adopts embarrassingly simple operator, pooling, to be the token mixer, and discovers that \textit{PoolFormer} effectively outperforms the delicate ResNet/ViT/MLP-like baselines~\cite{resnet, rsb, vit, deit, pvt, mlp-mixer, resmlp, liu2021pay}, which confirms the significance of MetaFormer.

In this paper, we make further steps exploring the boundaries of MetaFormer, through, again, deliberately taking our eyes off the token mixers. Our goal is to push the limits of MetaFomer, based on which we may have a comprehensive picture of its capacity. To this end, we adopt the most basic or common token mixers, and study the performance of the resultant MetaFormer models on the large-scale ImageNet-1K image classification. Specifically, we examine the token mixers being bare operators such as identity mapping or global random mixing, and being the common techniques dated back years ago such as separable convolution~\cite{chollet2017xception, mamalet2012simplifying, mobilenetv2} and vanilla self-attention~\cite{transformer}, as shown in Figure~\ref{fig:overall_framework}. We summarize our key experimental results in Figure \ref{fig:first_figure}, alongside our main observations are as follows.

\begin{itemize}
    \item \textbf{MetaFormer secures solid lower bound of performance}. 
    By specifying the token mixer to be the plainest operator, identity mapping, we build a MetaFormer model termed \textit{IdentityFormer} to probe the performance lower bound. This crude model, surprisingly, already achieves gratifying accuracy. For example, with 73M parameters and 11.5G MACs, IdentityFormer attains top-1 accuracy of 80.4\% on ImageNet-1K. Results of IdentityFormer demonstrate that MetaFormer is indeed a dependable architecture that ensures a favorable performance, even when the lowest degree of token mixing is involved. 
    \item \textbf{MetaFormer works well with arbitrary token mixers.}
    To explore MetaFormer's universality to token mixers, we further cast the token mixer to be random, with which the message passing between tokens is enabled but largely arbitrary. Specifically, we equip the token mixers with random mixing in the top two stages and preserve the identity mapping in the bottom two stages, to avoid bringing excessive computation cost and frozen parameters. The derived model, termed \textit{RandFormer}, turns out to be efficacious and improves IdentityFormer by 1.0\%, yielding an accuracy of 81.4\%. This result validates the MetaFormer's universal compatibility with token mixers. As such, please rest assured of MetaFormer's performance when exotic token mixers are introduced.
    \item \textbf{MetaFormer effortlessly offers state-of-the-art performance.}
    We make further attempts by injecting more informative operators into MetaFormer to probe its performance. Again, without introducing novel token mixers, MetaFormer models equipped with ``old-fashioned'' token mixers invented years ago, including inverted separable convolutions~\cite{chollet2017xception, mamalet2012simplifying, mobilenetv2} and vanilla self-attention~\cite{transformer}, readily delivers state-of-the-art results. Specifically,
    \begin{itemize}
     \item \textbf{ConvFormer outperforms ConvNeXt.}
      By instantiating the token mixer as separable depthwise convolutions, the resultant model, termed \textit{ConvFormer}, can be treated as a pure-CNN model without channel or spatial attention~\cite{senet, woo2018cbam, transformer, vit}. Experiments results showcase that ConvFormer consistently outperforms the strong pure-CNN model ConvNeXt~\cite{convnext}.
      \item \textbf{CAFormer sets new record on ImageNet-1K.}
      If we are to introduce attention into ConvFormer by even adopting the vanilla self-attention \cite{transformer}, the derived model, termed \textit{CAFormer}, readily yields record-setting performance on ImageNet-1K. Specifically, {CAFormer} replaces the token mixer of ConvFormer in the top two stages with vanilla self-attention, and hits a new record of 85.5\% top-1 accuracy at $224^2$ resolution on ImageNet-1K under the normal supervised setting (without extra data or distillation).
    \end{itemize}
\end{itemize}

These  MetaFormer models, with most basic or commonly-used token mixers, readily serve as dependable and competitive baselines for vision applications. When delicate token mixers or advanced training strategies are introduced, we will not be surprised at all to see the performance of MetaFormer-like models hitting new records.

Along our exploration, we also find that a new activation, \textit{StarReLU}, largely reduces the activation FLOPs up to 71\%, when compared with the commonly-adopted GELU. {StarReLU} is a variant of Squared ReLU, but particularly designed for alleviating distribution shifts. In our experiments, specifically, {StarReLU outperforms GELU by 0.3\%/0.2\% accuracy on ConvFormer-S18/CAFormer-S18, respectively.} We therefore expect {StarReLU} to find great potential in MetaFormer-like models alongside other neural networks.

\section{Method}
\subsection{Recap the concept of MetaFormer}
The concept MetaFormer \cite{metaformer} is a general architecture instead of a specific model, which is abstracted from Transformer \cite{transformer} by not specifying token mixer. Specifically, the input is first embedded as a sequence of features (or called tokens) \cite{transformer, vit}:
\begin{equation}
    X = \mathrm{InputEmbedding}(I).
\end{equation}
Then the token sequence $X \in \mathbb{R}^{N\times C}$ with length $N$ and channel dimension $C$ is fed into repeated MetaFormer blocks, one of which can be expressed as
\begin{align} 
X' & = X + \mathrm{TokenMixer}\left(\mathrm{Norm_1}(X)\right), \\
X'' &= X' + \sigma\left(\mathrm{Norm_2}(X')W_1\right)W_2,
\end{align}
where $\mathrm{Norm_1}(\cdot)$ and $\mathrm{Norm_2}(\cdot)$ are normalizations \cite{batch_norm, layer_norm}; $\mathrm{TokenMixer}(\cdot)$ means token mixer mainly for propagating information among tokens; $\sigma(\cdot)$ denotes activation function; $W_1$ and $W_2$ are learnable parameters in channel MLP. By specifying token mixers as concrete modules, MetaFormer is then instantiated into specific models.

\subsection{IdentityFormer and RandFormer}
Following \cite{metaformer}, we would like to instantiate token mixer as basic operators, to further probe the capacity of MetaFormer. The first one we considered is the identity mapping,
\begin{equation} 
\mathrm{IdentityMapping}(X) = X.
\end{equation}
Identity mapping does not conduct any token mixing, so actually, it can not be regarded as token mixer. For convenience, we still treat it as one type of token mixer to compare with other ones.

Another basic token mixer we utilize is global random mixing,
\begin{equation}
\label{eqn:random_mixing}
    \mathrm{RandomMixing}(X) = W_RX,
\end{equation}
where $X \in \mathbb{R}^{N \times C}$ is the input with sequence length $N$ and channel dimension $C$, and $W_R \in \mathbb{R}^{N \times N}$ is a matrix that are frozen after random initialization. This token mixer will bring extra frozen parameters and computation cost quadratic to the token number, so it is not suitable for large token number. 
The PyTorch-like code of the identity mapping and random mixing are shown in Algorithm \ref{alg:id_rand_conv}.

\begin{algorithm}[!ht]
\caption{Token mixers of identity mapping and random mixing, PyTorch-like Code}
\label{alg:id_rand_conv}
\definecolor{codeblue}{rgb}{0.25,0.5,0.5}
\definecolor{codekw}{rgb}{0.85, 0.18, 0.50}
\lstset{
  backgroundcolor=\color{white},
  basicstyle=\fontsize{7.5pt}{7.5pt}\ttfamily\selectfont,
  columns=fullflexible,
  breaklines=true,
  captionpos=b,
  commentstyle=\fontsize{7.5pt}{7.5pt}\color{codeblue},
  keywordstyle=\fontsize{7.5pt}{7.5pt}\color{codekw},
  showstringspaces=false,
  stringstyle=\color{darkred},
}
\begin{lstlisting}[language=python]
import torch
import torch.nn as nn

# Identity mapping
from torch.nn import Identity

# Random mixing
class RandomMixing(nn.Module):
    def __init__(self, num_tokens=196):
        super().__init__()
        self.random_matrix = nn.parameter.Parameter(
            data=torch.softmax(torch.rand(num_tokens, num_tokens), dim=-1),
            requires_grad=False)

    def forward(self, x):
        B, H, W, C = x.shape
        x = x.reshape(B, H*W, C)
        x = torch.einsum('mn, bnc -> bmc', self.random_matrix, x)
        x = x.reshape(B, H, W, C)
        return x
\end{lstlisting}
\end{algorithm}

To build the overall framework, we simply follow the 4-stage model \cite{alexnet, resnet} configurations of PoolFormer \cite{metaformer} to build models of different sizes. Specifically, we specify token mixer as identity mapping in all four stages and name the derived model IdentityFormer. To build RandFormer, considering that the token mixer of random mixing will bring much extra frozen parameters and computation cost for long token length, we thus remain identity mapping in the first two stages but set global random mixing as token mixer in the last two stages.

To compare IdentityFormer/RandFormer with PoolFormer \cite{metaformer} fairly, we also apply the techniques mentioned above to PoolFormer and name the new model PoolFormerV2. The model configurations are shown in Table \ref{tab:config_basic_models} and the overall frameworks are shown in Figure \ref{fig:overall_framework}.

\begin{table*}
\renewcommand{\arraystretch}{1.3}
  \caption{\textbf{Model configurations of IdentityFormer, RandFormer and PoolFormerV2.} ``C'', ``L'' and ``T'' means channel number, block number and token mixer type, respectively. ``Id'', "Rand'' and ``Pool'' denotes token mixer of identity mapping, random mixing and pooling, respectively. The contents in the tuples represent the configurations in the four stages of the models.}
  \label{tab:config_basic_models}
  \centering
  \begin{tabular}{l | l | c | c | c }
\whline
\multicolumn{2}{l|}{Model} & IdentityFormer & RandFormer  & PoolFormerV2 \\
\whline
\multirow{4}{*}{Size} & S12 & \multicolumn{3}{l}{\quad  \quad \quad \quad \quad $C = (64, 128, 320, 512)$, \quad  \quad \quad \quad  $L = (2, 2, 6, 2)$}\\
~ & S24 & \multicolumn{3}{l}{\quad  \quad \quad \quad \quad $C = (64, 128, 320, 512)$, \quad  \quad \quad \quad $L = (4, 4, 12, 4)$} \\
~ & S36 & \multicolumn{3}{l}{\quad  \quad \quad \quad \quad $C = (64, 128, 320, 512)$, \quad  \quad \quad \quad  $L = (6, 6, 18, 6)$} \\
~ & M36 & \multicolumn{3}{l}{\quad  \quad \quad \quad \quad $C = (96, 192, 384, 768)$, \quad  \quad \quad \quad  $L = (6, 6, 18, 6)$} \\
~ & M48 & \multicolumn{3}{l}{\quad  \quad \quad \quad \quad $C = (96, 192, 384, 768)$, \quad  \quad \quad \quad  $L = (8, 8, 24, 8)$} \\
\hline
\multicolumn{2}{l|}{Token Mixer}   & $T=(\mathrm{Id, Id, Id, Id})$ & $T= (\mathrm{Id, Id, Rand, Rand})$ & $T=(\mathrm{Pool, Pool, Pool, Pool})$ \\
\hline
\multicolumn{2}{l|}{Classifier Head} & \multicolumn{3}{c}{Global average pooling, Norm, FC} \\
\whline
\end{tabular}

\end{table*}

\begin{table*}
\renewcommand{\arraystretch}{1.3}
  \caption{
  \textbf{Model configurations of ConvFormer and CAFormer.} ``C'', ``L'' and ``T'' means channel number, block number and token mixer type. ``Conv'' and ``Attn'' denotes token mixer of separable convolution and vanilla self-attention, respectively. The contents in the tuples represent the configurations in the four stages of the models.}
  \label{tab:config_convformer_caformer}
\centering
\begin{tabular}{l | l | c | c}
\whline
\multicolumn{2}{l|}{Model} & ConvFormer & CAFormer \\
\whline
\multirow{4}{*}{Size} & S18 & \multicolumn{2}{l}{\quad \quad \quad$C = (64, 128, 320, 512)$, \quad  \quad \quad $L = (3, 3, 9, 3)$}\\
~ & S36 & \multicolumn{2}{l}{\quad \quad \quad$C = (64, 128, 320, 512)$, \quad  \quad \quad $L = (3, 12, 18, 3)$} \\
~ & M36 & \multicolumn{2}{l}{\quad \quad \quad$C = (96, 192, 384, 576)$, \quad  \quad \quad $L = (3, 12, 18, 3)$} \\
~ & B36 & \multicolumn{2}{l}{\quad \quad \quad$C = (128, 256, 512, 768)$, \quad  \quad \ \  $L = (3, 12, 18, 3)$} \\
\hline
\multicolumn{2}{l|}{Token Mixer}   & $T=(\mathrm{Conv, Conv, Conv, Conv})$ & $T= (\mathrm{Conv, Conv, Attn, Attn})$ \\
\hline
\multicolumn{2}{l|}{Classifier Head}   & \multicolumn{2}{c}{Global average pooling, Norm, MLP} \\
\whline
\end{tabular}

\end{table*}

\subsection{ConvFormer and CAFormer}
The above section utilizes basic token mixers to probe the lower bound of performance and model universality in terms of token mixers. In this section, without designing novel token mixers, we just specify the token mixer as commonly-used operators to probe the model potential for achieving state-of-the-art performance. The first token mixer we choose is depthwise separable convolution \cite{chollet2017xception, mamalet2012simplifying}. Specifically, we follow the inverted separable convolution module in MobileNetV2 \cite{mobilenetv2},
\begin{equation}
\label{eqn:convolutions}
    \mathrm{Convolutions(X)} = \mathrm{Conv_{pw2}}(\mathrm{Conv_{dw}}(\sigma(\mathrm{Conv_{pw1}}(X)))),
\end{equation}
where $\mathrm{Conv_{pw1}}(\cdot)$ and $\mathrm{Conv_{pw2}}(\cdot)$ are pointwise convolutions, $\mathrm{Conv_{dw}}(\cdot)$ is the depthwise convolution, and $\sigma(\cdot)$ means the non-linear activation function. 
In practice, we set the kernel size as 7 following \cite{convnext} and the expansion ratio as 2. We instantiate the MetaFormer as \textit{ConvFormer} by specifying the token mixers as the above separable convolutions. ConvFormer also adopts 4-stage framework \cite{resnet, pvt, swin} (Figure \ref{fig:overall_framework}) and the model configurations of different sizes are shown in the Table \ref{tab:config_convformer_caformer}.

Besides convolutions, another common token mixer is vanilla self-attention \cite{transformer} used in Transformer. This global operator is expected to have better ability to capture long-range dependency. However, since the computational complexity of self-attention is quadratic to the number of tokens, it will be cumbersome to adopt vanilla self-attention in the first two stages that have many tokens. As a comparison, convolution is a local operator with computational complexity linear to token length. Inspired by \cite{detr, vit, dai2021coatnet, metaformer}, we adopt 4-stage framework and specify token mixer as convolutions in the first two stages and attention in the last two stages to build \textit{CAFormer}, as shown in Figure \ref{fig:overall_framework}. See Table \ref{tab:config_convformer_caformer} for model configurations of different sizes.

\subsection{Techniques to improve MetaFormer}
This paper does not introduce complicated token mixers. Instead, we introduce a new activation StarReLU and other two modifications \cite{shleifer2021normformer, raffel2020exploring, chowdhery2022palm} to improve MetaFormer. 
\subsubsection{StarReLU}
In vanilla Transformer \cite{transformer}, ReLU \cite{relu} is chosen as the activation function  that can be expressed as
\begin{equation}
\label{eqn:relu}
    \mathrm{ReLU}(x) = \mathrm{max}(0, x),
\end{equation}
where $x$ denotes any one neural unit of the input. This activation costs 1 FLOP for each unit. Later, GPT \cite{gpt1} uses GELU \cite{gelu} as activation and then many subsequent Transformer models (\eg, BERT \cite{bert}, GPT-3 \cite{gpt3} and ViT \cite{vit}) employ this activation by default. GELU can be approximated as,
\begin{align}
&\mathrm{GELU}(x) = x\Phi(x) \\
&\approx 0.5 \times x(1 + \mathrm{tanh}(\sqrt{2 / \pi}(x + 0.044715 \times x^{3}))),
\label{eqn:gelu}
\end{align}
where $\Phi(\cdot)$ is the Cumulative Distribution Function for Gaussian Distribution (CDFGD).  
Although revealing better performance than ReLU \cite{so2021primer, metaformer}, GELU approximately brings 14 FLOPs \footnote{$\mathrm{tanh}$ is counted 6 FLOPs for simplicity \cite{tanh_flops}.}, much larger then ReLU's 1 FLOP of cost. To simplify GELU, \cite{so2021primer}  finds that CDFGD can be replaced by ReLU, 
\begin{equation}
\label{eqn:squared_relu}
    \mathrm{SquaredReLU}(x) = x\mathrm{ReLU}(x) = (\mathrm{ReLU}(x))^{2}.
\end{equation}
This activation is called Squared ReLU \cite{so2021primer}, only costing 2 FLOPs for each input unit. Despite the simplicity of Squared ReLU, we find its performance can not match that of GELU for some models on image classification task as shown in Section \ref{sec:ablation}. We hypothesize the worse performance may be resulted from the distribution shift of the output \cite{klambauer2017self}. Assuming input $x$ follows normal distribution with mean 0 and variance 1, \ie  $x \sim N(0, 1)$,  we have:
\begin{equation}
\label{eqn:expectation_variance}
    \mathrm{E}\left((\mathrm{ReLU}(x))^2\right) = 0.5,
    \quad
    \mathrm{Var}\left((\mathrm{ReLU}(x))^2\right) = 1.25.
\end{equation}
See the appendix for the derivation process of Equation \ref{eqn:expectation_variance}. Therefore the distribution shift can be solved by 
\begin{align}
    \mathrm{StarReLU}(x) &= \frac{\left(\mathrm{ReLU}(x)\right)^2 - \mathrm{E}\left((\mathrm{ReLU}(x))^2\right)}{\sqrt{\mathrm{Var}\left((\mathrm{ReLU}(x))^2\right)}} \\
    &= \frac{\left(\mathrm{ReLU}(x)\right)^2 - 0.5}{\sqrt{1.25}} \\
    &\approx 0.8944 \cdot \left(\mathrm{ReLU}(x)\right)^2 - 0.4472.
\end{align}
We name the above activation \textit{StarReLU} as multiplications (*) is heavily used. However, the assumption of standard normal distribution for input is strong \cite{klambauer2017self}. To make the activation adaptable to different situations, like different models or initialization, scale and bias can be set to be learnable \cite{he2015delving, chen2020dynamic}. We uniformly re-write the activation as
\begin{equation}
    \mathrm{StarReLU}(x) = s \cdot (\mathrm{ReLU}(x))^{2} + b,
\end{equation}
where $s\in \mathbb{R}$ and $ b \in \mathbb{R}$ are scalars of scale and bias respectively, which are shared for all channels and can be set to be constant or learnable to attain different StarReLU variants. StarReLU only costs 4 FLOPs (or 3 FLOPs with only $s$ or $b$), much fewer than GELU's 14 FLOPs but achieving better performance as shown in Section \ref{sec:ablation}. For convenience, \textit{we utilize StarReLU with learnable scale and bias as default activation in this paper} as intuitively this variant can more widely adapt to different situations \cite{he2015delving, chen2020dynamic}. We leave the study of StarReLU variant selection for different situations in the future.

\subsubsection{Other modifications}
\myPara{Scaling branch output}
To scale up Transformer model size from depth, \cite{cait} proposes \textit{LayerScale} that multiplies layer output by a learnable vector:
\begin{equation}
    X' = X + \lambda_l \odot \mathcal{F}(\mathrm{Norm}(X)),
\end{equation}
where $X \in \mathbb{R}^{N \times C}$ denotes the input features with sequence length $N$ and channel dimension $C$, $\mathrm{Norm}(\cdot)$ is the normalization, $\mathcal{F}(\cdot)$ means the token mixer or channel MLP module, $\lambda_l \in \mathbb{R}^{C}$ represents the learnable LayerScale parameters initialized by a small value like 1e-5, and $\odot$ means element multiplication. 
Similar to LayerScale, \cite{zhu2021gradinit, liu2020understanding, shleifer2021normformer} attempt to stabilize architectures by scaling the residual branch (\textit{ResScale} \cite{shleifer2021normformer}):
\begin{equation}
    X' = \lambda_r \odot X + \mathcal{F}(\mathrm{Norm}(X)),
\end{equation}
where $\lambda_r \in \mathbb{R}^{C}$ denotes learnable parameters initialized as 1. 
Apparently, we can merge the above two techniques into \textit{BranchScale} by scaling all branches:
\begin{equation}
    X' = \lambda_r \odot X + \lambda_l \odot \mathcal{F}(\mathrm{Norm}(X)).
\end{equation}
Among these three scaling techniques, we find ResScale performs best according to our experiments in Section \ref{sec:ablation}. Thus, \textit{we adopt ResScale \cite{shleifer2021normformer} by default in this paper.}

\myPara{Disabling biases} Following \cite{raffel2020exploring, chowdhery2022palm}, we disable the biases of fully-connected layers, convolutions (if have) and normalization in the MetaFormer blocks, finding it does not hurt performance and even can bring slight improvement for specific models as shown in the ablation study. For simplicity, \textit{we disable biases in MetaFormer blocks by default.}

\section{Experiments}
\begin{table}[t]
\renewcommand{\arraystretch}{1.3}
\caption{
\textbf{Performance on ImageNet-1K of RSB-ResNet and MetaFormer models with basic tokens of identity mapping, random maxing and pooling.} The underlined numbers mean the numbers of parameters that are frozen after random initialization.
}
\label{tab:basic_token_mixers_imagenet}
\centering
\begin{tabular}{l | c | c | c }
\whline
Model & Params (M) & MACs (G) & Top-1 (\%) \\
\whline 
RSB-ResNet-18 \cite{resnet, rsb} & 11.7 & 1.8 & 70.6 \\
IdentityFormer-S12 & 11.9 & 1.8 & 74.6 \\
RandFormer-S12 & 11.9 + \underline{0.2}  & 1.9 & 76.6 \\
PoolFormerV2-S12 \cite{metaformer} & 11.9 & 1.8 & 78.0 \\
\hline 
RSB-ResNet-34 \cite{resnet, rsb} & 21.8 & 3.7 & 75.5 \\
IdentityFormer-S24 & 21.3 & 3.4 & 78.2 \\
RandFormer-S24 & 21.3 + \underline{0.5} & 3.5 & 78.8 \\
PoolFormerV2-S24 \cite{metaformer} & 21.3 & 3.4 & 80.7 \\
\hline
RSB-ResNet-50 \cite{resnet, rsb} & 25.6 & 4.1 & 79.8 \\
IdentityFormer-S36 & 30.8 & 5.0 & 79.3 \\
RandFormer-S36 & 30.8 + \underline{0.7} & 5.2 & 79.5 \\
PoolFormerV2-S36 \cite{metaformer} & 30.8 & 5.0 & 81.6 \\
\hline
RSB-ResNet-101 \cite{resnet, rsb} & 44.5 & 7.9 & 81.3 \\
IdentityFormer-M36 & 56.1 & 8.8 & 80.0 \\
RandFormer-M36 & 56.1 + \underline{0.7} & 9.0 & 81.2 \\
PoolFormerV2-M36 \cite{metaformer} & 56.1 & 8.8 & 82.2 \\
\hline
RSB-ResNet-152 \cite{resnet, rsb} & 60.2 & 11.6 & 81.8 \\
IdentityFormer-M48 & 73.3 & 11.5 & 80.4 \\
RandFormer-M48 & 73.3 + \underline{0.9} & 11.9 & 81.4 \\
PoolFormerV2-M48 \cite{metaformer} & 73.3 & 11.5 & 82.6 \\
\whline
\end{tabular}

\end{table}

\begin{table}[h]
\caption{
\textbf{Comparison among ViT (DeiT), isotropic IdentityFormer and isotropic IdentityFormer with stem of 4 convolutional layers} with stride of 2 and kernel size of $7^2$, $3^2$, $3^2$, and $3^2$ respectively.
}
\label{tab:iso}
\addtolength{\tabcolsep}{-2pt}
\centering
\small
\begin{tabular}{l | c c c }
\whline
\multirow{2}{*}{\makecell[c]{Model}} &  \multirow{2}{*}{\makecell[c]{Params \\ (M)}} & \multirow{2}{*}{\makecell[c]{MACs \\ (G)}}  & \multirow{2}{*}{\makecell[c]{Top-1 \\ (\%)}} \\
~ & ~ & ~ &  \\
\whline
DeiT-S \cite{deit} & 22 & 4.6 & 79.8 \\ 
IdentityFormer-S (\textit{iso.}) & 22 & 4.2 & 68.2  \\ 
IdentityFormer-S (\textit{iso.}, conv stem) & 23 & 4.6 & 75.4 \\
\whline
\end{tabular}
\end{table}

\begin{table*}[h!]
\renewcommand{\arraystretch}{1.3}
    \caption{
    \textbf{Performance of models trained on ImageNet-1K at the resolution of $224^2$ and finetuned at $384^2$.} Model highlighted with \colorbox{mygray}{gray background} are proposed in this paper. The column ``MetaFormer'' denotes whether models adopt MetaFormer architecture (partially). * To the best of our knowledge, the model \textbf{sets a new record on ImageNet-1K} with the accuracy of 85.5\% at $224^2$ resolution under normal supervised setting (without external data or distillation), surpassing the previous best record of 85.3\% set by MViTv2-L \cite{li2022mvitv2} with 55\% fewer parameters and 45\% fewer MACs.}
    \label{tab:imagenet}
    \centering
    \begin{tabular}{l | c | c | c | c c | c c }
\whline
\multirow{3}{*}{\makecell[c]{Model}}   &  \multirow{3}{*}{\makecell[c]{MetaFormer}} & \multirow{3}{*}{\makecell[c]{Mixing Type}}     & \multirow{3}{*}{\makecell[c]{Params (M)}}   & \multicolumn{4}{c}{Testing at resolution} \\
\cline{5-8}
~ & ~ & ~ & ~ &  \multicolumn{2}{c|}{@224} & \multicolumn{2}{c}{$\uparrow$384} \\ 
~ & ~ & ~ & ~ & MACs (G) & Top-1 (\%) & MACs (G)  & Top-1 (\%) \\
\whline
RSB-ResNet-50 \cite{resnet, rsb} & \xmarkg & Conv   & 26 & 4.1 & 79.8 & - & - \\
RegNetY-4G \cite{regnet, rsb} & \xmarkg & Conv   & 21 & 4.0 & 81.3 & - & - \\
ConvNeXt-T \cite{convnext} & \xmarkg & Conv   & 29 & 4.5 & 82.1 & - & - \\
VAN-B2 \cite{guo2022visual} & \cmark & Conv & 27 & 5.0 & 82.8 & - & - \\
\gr
ConvFormer-S18 & \cmark & Conv  & 27 & 3.9 & \textbf{83.0} & 11.6 & 84.4 \\
\hdashline
DeiT-S \cite{deit} & \cmark & Attn  & 22 & 4.6 & 79.8 & - & - \\
T2T-ViT-14 \cite{t2t} & \cmark & Attn   & 22 & 4.8 & 81.5 & 17.1 & 83.3 \\
Swin-T \cite{swin} & \cmark & Attn   & 29 & 4.5 & 81.3 & - & - \\
CSWin-T \cite{dong2022cswin} & \cmark & Attn  & 23 & 4.3 & 82.7 & - & - \\
MViTv2-T \cite{li2022mvitv2} & \cmark & Attn & 24 & 4.7 & 82.3 & - & - \\
\textcolor{black}{Dual-ViT-S \cite{yao2022dual}} & \cmark & \textcolor{black}{Attn} & \textcolor{black}{25} & \textcolor{black}{4.8} & \textcolor{black}{83.4} & - & - \\
CoAtNet-0 \cite{dai2021coatnet} & \cmark & Conv + Attn  & 25 & 4.2 & 81.6 & 13.4 & 83.9 \\
UniFormer-S \cite{li2022uniformer} & \cmark & Conv + Attn  & 22 & 3.6 & 82.9 & - & - \\
iFormer-S \cite{si2022inception} & \cmark & Conv + Attn & 20 & 4.8 & 83.4 & 16.1 & 84.6 \\
\gr
CAFormer-S18 & \cmark & Conv + Attn  & 26 & 4.1 & \textbf{83.6} & 13.4 & 85.0 \\
\whline
RSB-ResNet-101 \cite{resnet, rsb} & \xmarkg & Conv   & 45 & 7.9 & 81.3 & - & - \\
RegNetY-8G \cite{regnet, rsb} & \xmarkg & Conv   & 39 & 8.0 & 82.1 & - & - \\
ConvNeXt-S \cite{convnext} & \xmarkg & Conv   & 50 & 8.7 & 83.1 & - & - \\
VAN-B3 \cite{guo2022visual} & \cmark & Conv & 45 & 9.0 & 83.9 & - & - \\
\gr
ConvFormer-S36 & \cmark & Conv  & 40 & 7.6 & \textbf{84.1} & 22.4 & 85.4 \\
\hdashline
T2T-ViT-19 \cite{t2t} & \cmark & Attn   & 39 & 8.5 & 81.9 & - & - \\
Swin-S \cite{swin} & \cmark & Attn   & 50 & 8.7 & 83.0 & - & - \\
CSWin-S \cite{dong2022cswin} & \cmark & Attn  & 35 & 6.9 & 83.6 & - & - \\
MViTv2-S \cite{li2022mvitv2} & \cmark & Attn & 35 & 7.0 & 83.6 & - & - \\
CoAtNet-1 \cite{dai2021coatnet} & \cmark & Conv + Attn  & 42 & 8.4 & 83.3 & 27.4 & 85.1 \\
UniFormer-B \cite{li2022uniformer} & \cmark & Conv + Attn  & 50 & 8.3 & 83.9 & - & - \\
\gr
CAFormer-S36 & \cmark & Conv + Attn  & 39 & 8.0 & \textbf{84.5} & 26.0 & 85.7 \\
\whline
RSB-ResNet-152 \cite{resnet, rsb} & \xmarkg & Conv   & 60 & 11.6 & 81.8 & - & - \\
RegNetY-16G \cite{regnet, rsb} & \xmarkg & Conv   & 84 & 15.9 & 82.2 & - & - \\
ConvNeXt-B \cite{convnext} & \xmarkg & Conv   & 89 & 15.4 & 83.8 & 45.0 & 85.1 \\
RepLKNet-31B \cite{ding2022scaling} & \cmark & Conv & 79 & 15.3 & 83.5 & 45.1 & 84.8 \\
VAN-B4 \cite{guo2022visual} & \cmark & Conv & 60 & 12.2 & 84.2 & - & - \\
SLaK-B \cite{liu2022more} & \cmark & Conv & 95 & 17.1 & 84.0 & 50.3 & 85.5 \\
\gr
ConvFormer-M36 & \cmark & Conv  & 57 & 12.8 & \textbf{84.5} & 37.7 & 85.6\\
\hdashline
DeiT-B \cite{deit} & \cmark & Attn   & 86 & 17.5 & 81.8 & 55.4 & 83.1 \\
T2T-ViT-24 \cite{t2t} & \cmark & Attn   & 64 & 13.8 & 82.3 & - & - \\
Swin-B \cite{swin} & \cmark & Attn   & 88 & 15.4 & 83.5 & 47.1 & 84.5 \\
CSwin-B \cite{dong2022cswin} & \cmark & Attn   & 78 & 15.0 & 84.2 & 47.0 & 85.4 \\
MViTv2-B \cite{li2022mvitv2} & \cmark & Attn & 52 & 10.2 & 84.4 & 36.7 & 85.6 \\
CoAtNet-2 \cite{dai2021coatnet} & \cmark & Conv + Attn  & 75 & 15.7 & 84.1 & 49.8 & 85.7 \\
MaxViT-S \cite{tu2022maxvit} & \cmark & Conv + Attn  & 69 & 11.7 & 84.5 & 36.1 & 85.2 \\
iFormer-L \cite{si2022inception} & \cmark & Conv + Attn & 87 & 14.0 & 84.8 & 45.3 & 85.8 \\
\gr
CAFormer-M36 & \cmark & Conv + Attn  & 56 & 13.2 & \textbf{85.2} & 42.0 & 86.2 \\
\whline
RegNetY-32G \cite{regnet, rsb} & \xmarkg & Conv   & 145 & 32.3 & 82.4 & - & - \\
ConvNeXt-L \cite{convnext} & \xmarkg & Conv   & 198 & 34.4 & 84.3 & 101.0 & 85.5 \\
\gr
ConvFormer-B36 & \cmark & Conv  & 100 & 22.6 & \textbf{84.8} & 66.5 & 85.7 \\
\hdashline
MViTv2-L \cite{li2022mvitv2} & \cmark & Attn & 218 & 42.1 & 85.3 & 140.2 & 86.3 \\
CoAtNet-3 \cite{dai2021coatnet} & \cmark & Conv + Attn  & 168 & 34.7 & 84.5 & 107.4 & 85.8 \\
MaxViT-B \cite{tu2022maxvit} & \cmark & Conv + Attn  & 120 & 23.4 & 85.0 & 74.2 & 86.3 \\
\gr
CAFormer-B36 & \cmark & Conv + Attn  & 99 & 23.2 & \textbf{85.5}* & 72.2 & 86.4 \\
\whline
\end{tabular}

\end{table*}

\subsection{Image Classification}
\subsubsection{Setup} 
ImageNet-1K \cite{imagenet} image classification is utilized to benchmark these baseline models. ImageNet-1K is one of the most widely-used datasets in computer vision which contains about 1.3M images of 1K classes on training set, and 50K images on validation set. For ConvFormer-B36 and CAFormer-B36, we also conduct pre-training on ImageNet-21K \cite{imagenet, imagenet_ijcv}, a much larger dataset containing $\sim$14M images of 21841 classes, and then fine-tune the pretrained model on ImageNet-1K for evaluation. Our implementation is based on PyTorch library \cite{paszke2019pytorch} and \texttt{Timm} codebase \cite{rw2019timm} and the experiments are run on TPUs.

\myPara{Training and fine-tuning on ImageNet-1K}
We mainly follow the hyper-parameters of DeiT \cite{deit}. Specifically, models are trained for 300 epochs at $224^2$ resolution. Data augmentation and regularization techniques include RandAugment \cite{cubuk2020randaugment}, Mixup \cite{zhang2017mixup}, CutMix \cite{yun2019cutmix}, Random Erasing \cite{zhong2020random}, weight decay, Label Smoothing \cite{szegedy2016rethinking} and Stochastic Depth \cite{huang2016deep}. We do not use repeated augmentation \cite{berman2019multigrain, hoffer2020augment} and LayerScale \cite{cait}, but use ResScale \cite{shleifer2021normformer} for the last two stages. We adopt AdamW \cite{kingma2014adam, loshchilov2017decoupled} optimizer with batch size of 4096 for most models except CAFormer since 
we found it suffers a slight performance drop compared with that with batch size of 1024.
The problem may be caused by the large batch size, so we use a large-batch-size-friendly optimizer LAMB \cite{you2019large} for CAFormer. For $384^2$ resolution, we fine-tune the models trained at $224^2$
resolution for 30 epochs with Exponential Moving Average (EMA) \cite{ema}. The details of hyper-parameters are shown in the appendix. 

\myPara{Pre-training on ImageNet-21K and fine-tuning on ImageNet-1K} To probe the scaling capacity with a larger dataset, we pre-train ConvFormer and CAFormer on ImageNet-21K for 90 epochs at the resolution of $224^2$. Then the pre-trained models are fine-tuned on ImageNet-1K at the resolution of $224^2$ and $384^2$ for 30 epochs with EMA \cite{ema}. See the appendix for more details of hyper-parameters.

\myPara{Robustness evaluation}
Following ConvNeXt \cite{convnext}, we also directly evaluate our
ImageNet models on several robustness benchmarks, \ie~ImageNet-C \cite{imagenet-c}, ImageNet-A \cite{imagenet-a}, ImageNet-R \cite{imagenet-r} and ImageNet-Sketch \cite{imagenet-sketch}. Note that we do not adopt additional fine-tuning or any specialized modules. Mean corruption error
(mCE) is reported for ImageNet-C and top-1 accuracy is for all other datasets.

\subsubsection{Results of Models with basic token mixers}
Table \ref{tab:basic_token_mixers_imagenet} shows the performance of models with basic token mixers on ImageNet-1K. Surprisingly, with bare identity mapping as token mixer, IdentityFormer already performs very well, especially for small model sizes. For example, IdentityFormer-S12/S24 outperforms RSB-ResNet-18/34 \cite{resnet, rsb} by 4.0\%/2.7\%, respectively. We further scale up the model size of IdentityFormer to see what accuracy it can achieve. By scaling up model size to $\sim$73M parameters and $\sim$12G MACs, IdentityFormer-M48 can achieve accuracy of 80.4\%. Without considering the comparability of model size, this accuracy already surpasses 79.8\% of RSB-ResNet-50. The results of IdentityFormer indicate that MetaFormer ensures solid lower bound of performance. That is to say, if you adopt MetaFormer as general framework to develop your own models, the accuracy will not be below 80\% with similar parameter numbers and MACs of IdentityFormer-M48.

To see whether the amazing performance of IdentityFormer is from hierarchical structure, we follow ViT-S (DeiT-S) to build isotropic IdentityFormer, and the results are shown in Table \ref{tab:iso}. IdentityFormer-S (\textit{iso.}) can achieve 68.2\% accuracy and IdentityFormer-S (\textit{iso.}) with the stem of 4 convolutional layers can even obtain an accuracy of 75.4\%. These results show that isotropic IdentityFormer also works well, demonstrating the performance of IdentityFormer is not from hierarchical structure. An important factor for model performance is the receptive field of the stem or downsampling layers, based on the large improvement of IdentityFormer-S (\textit{iso.}, conv stem) over IdentityFormer-S (\textit{iso.}).

Another surprising finding is that by replacing token mixer of IdentityFormer with random mixing in the top two stages, RandFormer can consistently improve IdentityFormer. For example, RandFormer-S12/M48 obtains accuracy of 76.6\%/81.4\%, surpassing IdentityFormer-S12/M48 by 2.0\%/1.0\%, respectively. For medium and large model sizes, RandFormer can also achieve accuracy comparable to RSB-ResNet, like RandFormer-M36's 81.2\% \vs RSB-ResNet-101's 81.3\%. The promising performance of RandFormer, especially its consistent improvement over IdentityFormer, demonstrates MetaFormer can work well with arbitrary token mixers and validates MetaFormer's universal compatibility with token mixers. Therefore, rest assured of MetaFormer's performance when exotic token mixers are equipped.

Compared with PoolFormerV2 \cite{metaformer} with basic token mixer of pooling, neither of IdentityFormer nor RandFormer can match its performance. The worse performance of IdentityFormer makes sense as identity mapping does not conduct any token mixing. The performance gap between RandFormer and PoolFormerV2 may result from the local inductive bias of pooling.

\subsubsection{Results of models with commonly-used token mixers}
We build ConvFormer by specifying the token mixer in MetaFormer as separable convolutions \cite{chollet2017xception, mamalet2012simplifying} used in MobileNetV2 \cite{mobilenetv2}. Meanwhile, CAFormer is built with token mixers of separable convolutions in the bottom two stages and vanilla self-attention in the top two stages. The results of models trained on ImageNet-1K are shown in Table \ref{tab:imagenet}. 

ConvFormer actually can be regarded as pure CNN-based model without any attention mechanism \cite{senet, woo2018cbam, transformer, vit}. It can be observed that ConvFormer outperforms strong CNN model ConvNeXt \cite{convnext} significantly. For example, at the resolution of $224^2$, ConvFomer-B36 (100M parameters and 22.6G MACs) surpasses ConvNeXt-B (198M parameters and 34.4G MACs) by 0.5\% top-1 accuracy while only requiring 51\% parameters and 66\% MACs. 
Compared with another stong CNN model EfficientNetV2-L \cite{tan2021efficientnetv2} with input size of $480^2$ (120M parameters, 53.0G MACs, 85.7\% top-1 accuracy), ConvFormer-B36 with input size of $384^2$ can match its accuracy.

Also, ConvFormer outperforms various strong attention-based or hybrid models. For instance, ConvFormer-M36 outperforms Swin-B \cite{swin}/CoAtNet-2 \cite{dai2021coatnet} by 1.0\%/0.4\% with 35\%/24\% fewer parameters and 17\%/18\% fewer MACs. 

\begin{figure*}[t]
  \centering
   \includegraphics[width=1\linewidth]{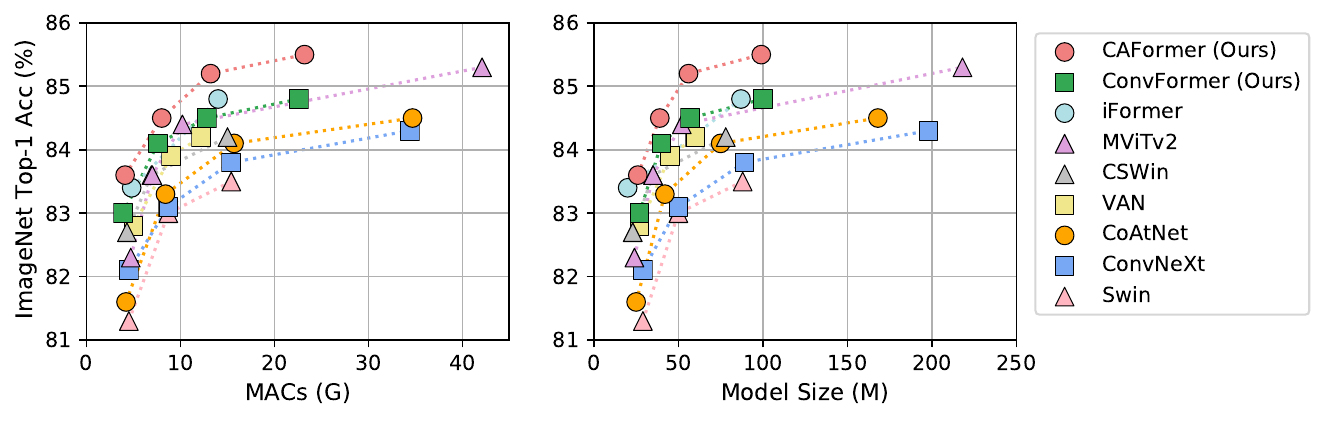}
   \caption{\textbf{ImageNet-1K validation accuracy \vs MACs/Model Size at the resolution of $224^2$.} Models with token (feature) mixing based on convolution, attention or hybrid are presented by $\square$, $\triangle$ or $\iscircle$ respectively.} 
   \label{fig:overall_comparision}
\end{figure*}

Besides ConvFormer, CAFormer achieves more remarkable performance. Although CAFormer is just built by equipping token mixers of separable convolutions \cite{chollet2017xception, mamalet2012simplifying, mobilenetv2} in bottom stages and vanilla self-attention \cite{transformer} in top stages, it already consistently outperforms other models in different sizes, as clearly shown in Figure \ref{fig:overall_comparision}. Remarkably, to the best of our knowledge, CAFormer \textbf{sets new record on ImageNet-1K} with top-1 accuracy of 85.5\% at $224^2$ resolution under normal supervised setting (without external data or distillation models).

When pre-trained on ImageNet-21K (Table \ref{tab:in21k}), the model performance is further improved. For instance, the performance of ConvFormer-B36 and CAFormer-B36 surges to 87.0\% and 87.4\%, with 2.2\% and 1.9\% accuracy improvement compared with the results of ImageNet-1K training only. Both models keep superior to Swin-L \cite{swin}/ConvNeXt-L \cite{convnext}, showing the promising scaling capacity with a larger pre-training dataset. For example, ConvFormer-B36 outperforms ConvNeXt-L by 0.2\% with 49\% fewer parameters and 34\% fewer MACs. Compared with another strong model EfficientNetV2-XL (input size of $480^2$, 94.0G MACs, 208M parameters, 87.3\% top-1 accuracy), ConvFormer-B36 and CAForemer-B36 surpass it by 0.3\% and 0.8\% with only half of the parameters and 71\%/77\% MACs, respectively.

Just equipped with ``old-fashioned'' token mixers, ConvFormer and CAFormer instantiated from MetaFormer already can achieve remarkable performance, especially CAFormer sets a new record on ImageNet-1K. These results demonstrate MetaFormer can offer high potential for achieving state-of-the-art performance. When advanced token mixers or training strategies are introduced, we will not be surprised to see the performance of MetaFormer-like models setting new records. We expect ConvFormer and CAFormer as well as IdentityFormer and RandFormer to be dependable baselines for future neural architecture design.

\subsubsection{Robustness of models with commonly-used token mixers}

The robustness results of ConvFormer, CAFormer and other SOTA models are shown in Table \ref{tab:robustness}. Compared with Swin \cite{swin} and ConvNeXt \cite{convnext}, ConvFormer exhibits better or competitive performance. For example, for models trained on ImageNet-1K, ConvFormer-S18 obtains 25.3\% and 48.7\% on ImageNet-A \cite{imagenet-a} and ImageNet-R \cite{imagenet-r}, outperforming ConvNeXt-T by 1.1\% and 1.5\%, respectively. CAFormer attains more impressive performance: It not only consistently outperforms Swin and ConvNeXt, but also surpasses SOTA robust method FAN \cite{zhou2022understanding}. For instance, for models trained on ImageNet-1K, CAFormer-M36/CAFormer-B36 obtain 45.6\% and 48.5\% on ImageNet-A \cite{imagenet-a}, surpassing FAN \cite{zhou2022understanding} by 8.4\%/11.3\%, respectively.

\begin{table*}[h!]
\renewcommand{\arraystretch}{1.3}
    \caption{
    \textbf{Model evaluation of robustness.} Model highlighted with \colorbox{mygray}{gray background} are proposed in this paper. Note that we do not adopt additional fine-tuning or any specialized modules.
    }
    \label{tab:robustness}
    \centering
    \begin{tabular}{l | c | c c c c c c | c c c c c c}
\whline
\multirow{4}{*}{\makecell[c]{Model}}   &  \multirow{4}{*}{\makecell[c]{Params \\ (M)}}   & \multicolumn{12}{c}{Testing at resolution} \\
\cline{3-14}
~ & ~ &  \multicolumn{6}{c|}{@224} & \multicolumn{6}{c}{$\uparrow$384} \\ 
~ & ~ & \multirow{2}{*}{\makecell[c]{MACs \\ (G)}}  & \multirow{2}{*}{\makecell[c]{Clean \\ (\%)}} & \multirow{2}{*}{\makecell[c]{C ($\downarrow$) \\ (mCE)}} & \multirow{2}{*}{\makecell[c]{A \\ (\%)}} &  \multirow{2}{*}{\makecell[c]{R \\ (\%)}} &  \multirow{2}{*}{\makecell[c]{SK \\ (\%)}} & \multirow{2}{*}{\makecell[c]{MACs \\ (G)}}  & \multirow{2}{*}{\makecell[c]{Clean \\ (\%)}} & \multirow{2}{*}{\makecell[c]{C ($\downarrow$) \\ (mCE)}} & \multirow{2}{*}{\makecell[c]{A \\ (\%)}} &  \multirow{2}{*}{\makecell[c]{R \\ (\%)}} &  \multirow{2}{*}{\makecell[c]{SK \\ (\%)}} \\
~ & ~ & ~ & ~ & ~ & ~ & ~ & ~ & ~ & ~ & ~ & ~ & ~ & ~ \\
\whline
\multicolumn{14}{c}{Trained on ImageNet-1K} \\
Swin-T \cite{swin} &  29 & 4.5 & 81.3 & 62.0 & 21.6 & 41.3 & 29.1 & - & - & - & - & - & - \\
RVT-S* \cite{mao2022towards} & 23 & 4.7 & 81.9 & 49.4 & 25.7 & 47.7 & 34.7 & - & - & - & - & - & - \\
ConvNeXt-T \cite{convnext}    & 29 & 4.5 & 82.1 & 53.2 & 24.2 & 47.2 & 33.8 & - & - & - & - & - & - \\
FAN-S \cite{zhou2022understanding} & 28 & 5.3 & 82.5 & 47.7 & 29.1 & 50.4 & - & - & - & - & - & - & - \\
\gr
ConvFormer-S18 &  27 & 3.9 & 83.0 & 51.7 & 25.3 & 48.7 & 35.2 & 11.6 & 84.4 & 51.0 & 42.0 & 50.7 & 36.2 \\
\gr
CAFormer-S18 &  26 & 4.1 & 83.6 & 47.4 & 33.5 & 48.7 & 36.6 & 13.4 & 85.0 & 46.1 & 48.9 & 51.3 & 37.7 \\
\hline
Swin-S \cite{swin} &  50 & 8.7 & 83.0 & 52.7 & 32.3 & 45.1 & 32.4 & - & - & - & - & - & - \\
ConvNeXt-S \cite{convnext} & 50 & 8.7 & 83.1 & 51.2 & 31.2 & 49.5 & 37.1 & - & - & - & - & - & - \\
FAN-B \cite{zhou2022understanding} & 54 & 10.4 & 83.6 & 44.4 & 35.4 & 51.8 & - & - & - & - & - & - & - \\
\gr
ConvFormer-S36 &  40 & 7.6 & 84.1 & 47.1 & 33.2 & 50.8 & 38.4 & 22.4 & 85.4 & 47.7 & 49.9 & 51.9 & 37.8 \\
\gr
CAFormer-S36 &  39 & 8.0 & 84.5 &  44.7  &  40.9  &  51.7  &  39.5 & 26.0 & 85.7 &  42.7  &  57.1  &  54.5  &  41.7 \\
\hline
Swin-B \cite{swin} &  88 & 15.4 & 83.5 & 54.4 & 35.8 & 46.6 & 32.4 & 47.1 & 84.5 & 49.4 & 45.3 & 47.0 & 32.9 \\
RVT-B* \cite{mao2022towards} & 92 & 17.7 & 82.6 & 46.8 & 28.5 & 48.7 & 36.0 & - & - & - & - & - & - \\
ConvNeXt-B \cite{convnext}    & 89 & 15.4 & 83.8 & 46.8 & 36.7 & 51.3 & 38.2 & 45.0 & 85.1 & 48.6 & 47.6 & 52.2 & 38.5 \\
FAN-L \cite{zhou2022understanding} & 81 & 15.8 & 83.9 & 43.3 & 37.2 & \textbf{53.1} & - & - & - & - & - & - & - \\
Robust-ResNet \cite{wang2023can} & - & 17.3 & 81.6 & \textbf{34.9} & - & 51.1 & 38.1 & - & - & - & - & - & - \\
\gr
ConvFormer-M36 &  57 & 12.8 & 84.5 &  46.5  &  37.6  &  51.0  &  39.2 & 37.7 & 85.6 &  48.4  &  53.5  &  52.2  &  38.5 \\
\gr
CAFormer-M36 &  56 & 13.2 & 85.2 &  42.6  &  45.6  &  51.7  &  39.6 & 42.0 & 86.2 &  \textbf{41.7}  &  60.2  &  \textbf{55.0}  &  41.5 \\
\hline
ConvNeXt-L \cite{convnext}    & 198 & 34.4 & 84.3 & 46.6 & 41.1 & 53.4 & 40.1 & 101.0 & 85.5 & 46.8 & 52.5 & 53.6 & 39.9 \\
\gr
ConvFormer-B36 &  100 & 22.6 & 84.8 &  46.3  &  40.1  &  51.1  &  39.5 & 66.5 & 85.7 &  48.1  &  55.3  &  52.2  &  38.9 \\
\gr
CAFormer-B36 &  99 & 23.2 & \textbf{85.5} &  42.6  &  \textbf{48.5}  &  \textbf{53.9}  &  \textbf{42.5} & 72.2 & \textbf{86.4} &  42.8  &  \textbf{61.9}  &  \textbf{55.0}  &  \textbf{42.5} \\
\hline
\multicolumn{14}{c}{Pretrained on ImageNet-21K} \\
ConvNeXt-T \cite{convnext}    & 29 & 4.5 & 82.9 & 52.3 & 36.6 & 51.0 & 38.5 & 13.1 & 84.1 & 51.5 & 45.8 & 51.3 & 38.9 \\
\gr
ConvFormer-S18 &  27 & 3.9 & 83.7 &  47.5  &  33.4  &  53.4  &  40.3 & 11.6 & 85.0 &  47.2  &  50.1  &  55.0  &  41.6 \\
\gr
CAFormer-S18 &  26 & 4.1 & 84.1 &  44.8  &  43.3  &  54.1  &  41.2 & 13.4 & 85.4 &  43.3  &  58.3  &  55.9  &  42.0 \\
\hline
ConvNeXt-S \cite{convnext} & 50 & 8.7 & 84.6 & 45.6 & 45.1 & 57.3 & 43.6 & 25.5 & 85.8 & 44.2 & 57.0 & 59.1 & 45.8 \\
\gr
ConvFormer-S36 &  40 & 7.6 & 85.4 &  41.0  &  47.3  &  58.9  &  46.9 & 22.4 & 86.4 &  41.3  &  62.9  &  59.9  &  47.1 \\
\gr
CAFormer-S36 &  39 & 8.0 & 85.8 &  38.5  &  55.5  &  60.7  &  46.7 & 26.0 & 86.9 &  36.8  &  70.6  &  63  &  48.5 \\
\hline
Swin-B \cite{swin} &  88 & 15.4 & 85.2 & 42.0 & 51.7 & 59.3 & 45.5 & 47.1 & 86.4 & 37.8 & 65.3 & 63.0 & 48.5 \\
ConvNeXt-B \cite{convnext}    & 89 & 15.4 & 85.8 & 41.9 & 54.8 & 61.8 & 49.8 & 45.0 & 86.8 & 43.1 & 62.3 & 64.9 & 51.6 \\
\gr
ConvFormer-M36 &  57 & 12.8 & 86.1 &  38.4  &  56.1  &  60.9  &  49.1 & 37.7 & 86.9 &  39.0  &  68.5  &  61.8  &  49.1 \\
\gr
CAFormer-M36 &  56 & 13.2 & 86.6 &  35.2  &  60.9  &  63.4  &  49.7 & 42.0 & 87.5 &  33.9  &  73.9  &  65.3  &  51.0 \\
\hline
Swin-L \cite{swin} & 197 & 34.5 & 86.3 & 38.0 & 61.2 & 63.7 & 49.0 & 103.9 & 87.3 & 34.5 & 70.7 & 66.0 & 50.4 \\
ConvNeXt-L \cite{convnext}    & 198 & 34.4 & 86.8 & 38.3 & 60.5 & 63.9 & 49.9 & 101.0 & 87.5 & 40.2 & 65.5 & 66.7 & 52.8 \\
\gr
ConvFormer-B36 &  100 & 22.6 & 87.0 &  35.0  &  63.3  &  65.3  &  52.7 & 66.5 & 87.6 &  35.8  &  73.5  &  66.5  &  52.9 \\

\gr
CAFormer-B36 &  99 & 23.2 & \textbf{87.4} &  \textbf{31.8}  &  \textbf{69.4}  &  \textbf{68.3}  &  \textbf{52.8} & 72.2 & \textbf{88.1} &  \textbf{30.8}  &  \textbf{79.5}  &  \textbf{70.4}  &  \textbf{54.5} \\
\whline
\end{tabular}

\end{table*}

\begin{table*}[t]
\renewcommand{\arraystretch}{1.3}
    \caption{
    \textbf{Performance of models pre-trained on ImageNet-21K and fine-tuned on ImageNet-1K for evaluation.} Model highlighted with \colorbox{mygray}{gray background} are proposed in this paper.
    }
    \label{tab:in21k}
    \centering
    \begin{tabular}{l | c | c | c | c c | c c }
\whline
\multirow{3}{*}{\makecell[c]{Model}}   &  \multirow{3}{*}{\makecell[c]{MetaFormer}} & \multirow{3}{*}{\makecell[c]{Mixing Type}}     & \multirow{3}{*}{\makecell[c]{Params (M)}}   & \multicolumn{4}{c}{Testing at resolution} \\
\cline{5-8}
~ & ~ & ~ & ~ &  \multicolumn{2}{c|}{@224} & \multicolumn{2}{c}{$\uparrow$384} \\ 
~ & ~ & ~ & ~ & MACs (G) & Top-1 (\%) & MACs (G)  & Top-1 (\%) \\
\whline
ConvNeXt-T \cite{convnext} & \xmarkg & Conv & 29 & 4.5 & 82.9 & 13.1 & 84.1 \\
\gr
ConvFormer-S18 & \cmark & Conv & 27 & 3.9 & \textbf{83.7} & 11.6 & 85.0 \\
\hdashline
\gr
CAFormer-S18 & \cmark & Conv + Attn & 26 & 4.1 & \textbf{84.1} & 13.4 & 85.4 \\
\hline
ConvNeXt-S \cite{convnext} & \xmarkg & Conv & 50 & 8.7 & 84.6 & 25.5 & 85.8 \\
\gr
ConvFormer-S36 & \cmark & Conv & 40 & 7.6 & \textbf{85.4} & 22.4 & 86.4 \\
\hdashline
\gr
CAFormer-S36 & \cmark & Conv + Attn & 39 & 8.0 & \textbf{85.8} & 26.0 & 86.9 \\
\hline
ConvNeXt-B \cite{convnext} & \xmarkg & Conv & 89 & 15.4 & 85.8 & 45.1 & 86.8 \\
\gr
ConvFormer-M36 & \cmark & Conv  & 57 & 12.8 & \textbf{86.1} & 37.7 & 86.9\\
\hdashline
Swin-B \cite{swin} & \cmark & Attn   & 88 & 15.4 & 85.2 & 47.1 & 86.4 \\
\gr
CAFormer-M36 & \cmark & Conv + Attn  & 56 & 13.2 & \textbf{86.6} & 42.0 & 87.5 \\
\hline
ConvNeXt-L \cite{convnext} & \xmarkg & Conv & 198 & 34.4 & 86.8 & 101.0 & 87.5 \\
\gr
ConvFormer-B36 & \cmark & Conv & 100 & 22.6 & \textbf{87.0} & 66.5 & 87.6 \\
\hdashline
Swin-L \cite{swin} & \cmark & Attn & 197 & 34.5 & 86.3 & 103.9 & 87.3 \\
\gr
CAFormer-B36 & \cmark & Conv + Attn & 99 & 23.2 & \textbf{87.4} & 72.2 & 88.1 \\
\whline
\end{tabular}

\end{table*}

\begin{table*}[h]
\renewcommand{\arraystretch}{1.3}
\caption{\textbf{Ablation for ConvFormer-S18/CAFormer-S18 on ImageNet-1K.} $^*$ $\alpha$ and $\beta$ denote learnable scalars shared for all channels.}
\label{tab:ablation}
\centering
\setlength{\tabcolsep}{6pt}
\begin{tabular}{l|l|c c c c}
\whline
	\multirow{2}{*}{Ablation} & \multirow{2}{*}{Variant}  & \multicolumn{2}{c}{Top-1 (\%)}  \\
	& & ConvFormer-S18 & CAFormer-S18 \\
\whline
-- & Baseline &  83.0 & 83.6 \\
\hline
\multirow{3}{*}{\tabincell{l}{Activation \\ types}}  & StarReLU $\rightarrow$ ReLU \cite{relu} &  82.1 \textcolor{gray}{(-0.9)} & 82.9 \textcolor{gray}{(-0.7)} \\
             & StarReLU $\rightarrow$ Squared ReLU \cite{so2021primer} &  82.6 \textcolor{gray}{(-0.4)} & 83.4 \textcolor{gray}{(-0.2)} \\
             & StarReLU $\rightarrow$ GELU \cite{gelu} & 82.7 \textcolor{gray}{(-0.3)} & 83.4 \textcolor{gray}{(-0.2)} \\
\hline
\multirow{6}{*}{\tabincell{l}{StarReLU \\ variants}} & $\alpha \cdot (\mathrm{ReLU}(x))^2 + \beta^*$\\
             & \quad $\rightarrow$ $\alpha \cdot (\mathrm{ReLU}(x))^2$ &  82.6 \textcolor{gray}{(-0.4)} & 83.6 \textcolor{gray}{(-0.0)} \\
             & \quad $\rightarrow$  $(\mathrm{ReLU}(x))^2 + \beta $ &  83.0 \textcolor{gray}{(-0.0)} & 83.5 \textcolor{gray}{(-0.1)} \\
             & \quad $\rightarrow$  $1/\sqrt{1.25} \cdot (\mathrm{ReLU}(x))^2 -0.5/\sqrt{1.25}$ & 83.0 \textcolor{gray}{(-0.0)} & 83.5 \textcolor{gray}{(-0.1)} \\
             & \quad $\rightarrow$ $1/\sqrt{1.25} \cdot (\mathrm{ReLU}(x))^2$  & 82.6 \textcolor{gray}{(-0.4)} & 83.4 \textcolor{gray}{(-0.2)} \\
             & \quad $\rightarrow$ $(\mathrm{ReLU}(x))^2 -0.5$ & 83.0 \textcolor{gray}{(-0.0)} & 83.3 \textcolor{gray}{(-0.3)} \\

\hline
\multirow{3}{*}{\tabincell{l}{Branch output \\ scaling}}  & ResScale \cite{shleifer2021normformer} $\rightarrow$ None  & 82.8 \textcolor{gray}{(-0.2)} & 83.2 \textcolor{gray}{(-0.4)} \\
             & ResScale \cite{shleifer2021normformer} $\rightarrow$ LayerScale \cite{cait} & 82.8 \textcolor{gray}{(-0.2)} & 83.0 \textcolor{gray}{(-0.6)} \\
             & ResScale \cite{shleifer2021normformer} $\rightarrow$ BranchScale & 82.9 \textcolor{gray}{(-0.1)} & 83.3 \textcolor{gray}{(-0.3)} \\
\hline
\tabincell{l}{Biases in \\ each block}  & \tabincell{l}{Disable biases of Norm, FC and Conv \\ \quad $\rightarrow$ Enable biases} &  83.0 \textcolor{gray}{(-0.0)} & 83.5 \textcolor{gray}{(-0.1)} \\
\whline
\end{tabular}

\end{table*}

\subsection{Object detection and instance segmentation}
\subsubsection{Setup}

We evaluate ConvFormer and CAFormer on COCO \cite{coco} which contains 118K training images and 5K validation images. Following Swin \cite{swin} and ConvNeXt \cite{convnext}, we adopt ConvFormer and CAFormer pretrained on ImageNet-1K as the backbones for Mask R-CNN \cite{he2017mask} and Cascade Mask R-CNN \cite{cai2018cascade}.  Due to the large image resolution, we find adopting CAFormer will result in out-of-memory. To solve it, we limit attention of CAFormer in sliding window \cite{beltagy2020longformer, hassani2023neighborhood}. We also adopt AdamW multi-scale and $3\times$ schedule training setting, following Swin and ConvNeXt.

\subsubsection{Results}

Table \ref{tab:coco} shows the results of ConvFormer, CAFormer, and other two strong backbones Swin and ConvNeXt for COCO object detection and instance segmentation. The models with backbones of ConvFormer and CAFormer consistently outperform Swin and ConvNeXt. For example, Cascade Mask R-CNN with CAFormer-S18 as backbone largely surpasses that with Swin-T/ConvNeXt-T, \ie, 52.3 \vs 50.4/50.4 for box AP, and 45.2 \vs 43.7/43.7 for mask AP.

\subsection{Semantic segmentation}
\subsubsection{Setup}

We also evaluate ConvFormer and CAFormer on the ADE20K \cite{ade20k} for semantic segmentation task. ADE20K consists of 20K/2K images on the training/validation set, including 150 semantic categories. Following Swin and ConvNeXt, we equip ConvFormer and CAFormer as backbones for UperNet \cite{upernet}. All models are trained with AdamW optimizer \cite{kingma2014adam, loshchilov2017decoupled} and batch size of 16 for 160K iterations. 

\subsubsection{Results}

Table \ref{tab:upernet} shows the results of UperNet with different backbones. ConvFormer and CAFormer as backbones obtain better performance compared with Swin and ConvNeXt. For instance, the model with CAFormer-S18 obtains 48.9 mIoU, higher than those with Swin-T/ConvNeXt-T by 3.1/2.2.

\subsection{Ablation}
\label{sec:ablation}
This paper does not design novel token mixers but utilizes three techniques to MetaFormer. Therefore, we conduct ablation study for them, respectively. ConvFormer-S18 and CAFormer-S18 on ImageNet-1K are taken as the baselines. The results are shown in Table \ref{tab:ablation}. When the StarReLU is replaced with ReLU \cite{relu}, the performance of ConvFormer-S18/CAFormer significantly drops from 83.0\%/83.6\% to 82.1\%/82.9\%, respectively. When the activation is Squared ReLU \cite{so2021primer}, the performance is already satisfied. But for ConvFormer-18, it can not match that of GELU \cite{gelu}. As for the StarReLU, it not only can reduce 71\% activation FLOPs compared with GELU, but also achieves better performance with 0.3\%/0.2\% accuracy improvement for ConvFormer-S18/CAForemr-S18, respectively. This result expresses the promising application potential of StarReLU in MetaFormer-like models and other neural networks. We further observe the performance of different StarReLU variants on ConvFormer-S18. We adopt StarReLU with learnable scale and bias by default because it does not need to meet the assumption of standard normal distribution for input \cite{klambauer2017self} and can be conveniently applied for different models and initialization \cite{he2015delving, chen2020dynamic}. But for specific model ConvFormer-18, StarReLU with learnable or frozen bias is enough since it can already match the accuracy of the default StarReLU version. We leave the study of StarReLU variant selection in the future.

For other techniques, we find ResScale \cite{shleifer2021normformer} performs best among the branch output scaling techniques; disabling biases \cite{raffel2020exploring, chowdhery2022palm} in each block does not affect the performance for ConvFormer-S18 and can bring improvement of 0.1\% for CAFormer-S18. We thus employ ResScale and disabling biases of each block by default.

\begin{table}
\caption[caption]{
\textbf{Performance of object detection and instance segmentation on COCO } with Mask R-CNN and Cascade Mask R-CNN. The MACs are measured with input size of $800 \times 1333$ ($^*$ except $896 \times 896$ of MaxViT). The FPS are measured on NVIDIA V100 GPU.
\label{tab:coco}
}
\tablestyle{6pt}{1.1}
\addtolength{\tabcolsep}{-5.5pt}
\begin{tabular}{@{}lcccccccc@{}}
\whline
Backbone & MACs (G) & FPS & $\text{AP}^{\text{box}}$ & $\text{AP}^{\text{box}}_{50}$ & $\text{AP}^{\text{box}}_{75}$ & $\text{AP}^{\text{mask}}$ & $\text{AP}^{\text{mask}}_{\text{50}}$ & $\text{AP}^{\text{mask}}_{75}$  \\
\whline
\multicolumn{9}{c}{\scriptsize{Mask R-CNN 3$\times$ schedule}} \\
 Swin-T      & 267  & 19.0    & 46.0 & 68.1 & 50.3 & 41.6 & 65.1 & 44.9 \\
 ConvNeXt-T    & 262  & 22.1    & 46.2 & 67.9 & 50.8 & 41.7 & 65.0 & 44.9 \\
 \gr
 ConvFormer-S18 & 251 & 18.3 & 47.7 & 69.6 & 52.3 & 42.6 & 66.3 & 45.9 \\
 \gr
 CAFormer-S18 & 254 & 18.0 & \textbf{48.6} & 70.5 & 53.4 & \textbf{43.7} & 67.5 & 47.4 \\
\hline
\multicolumn{9}{c}{\scriptsize{Cascade Mask R-CNN 3$\times$ schedule}} \\
 MaxViT-T & 475$^*$ & - &  52.1 & 71.9 &  56.8 & 44.6 & 69.1 & 48.4 \\
 MaxViT-S & 595$^*$& - & 53.1 & 72.5 & 58.1 & 45.4 & 69.8 & 49.5 \\
 MaxViT-B & 856$^*$& - & 53.4 & 72.9 & 58.1 & 45.7 & 70.3 & 50.0 \\
\hdashline
 Swin-T               & 745  & 8.5     & 50.4 & 69.2 & 54.7 & 43.7 & 66.6 & 47.3 \\
 ConvNeXt-T             & 741  & 9.4    & 50.4 & 69.1 & 54.8 & 43.7 & 66.5 & 47.3 \\
  \gr
 ConvFormer-S18 & 729 & 8.7 & 51.5 & 70.7 & 55.8 & 44.6 & 67.8 & 48.2 \\
 \gr
 CAFormer-S18 & 733 & 8.7 & \textbf{52.3} & 71.3 & 56.9 & \textbf{45.2} & 68.6 & 48.8 \\
 \hline
 Swin-S               & 838  & 7.8    & 51.9	& 70.7	& 56.3	& 45.0	& 68.2	& 48.8 \\
 ConvNeXt-S             & 827 & 8.6   & 51.9 & 70.8 & 56.5 & 45.0 & 68.4 & 49.1 \\
 \gr
 ConvFormer-S36 & 805 & 7.4 & 52.5 & 71.1 & 57.0 & 45.2 & 68.6 & 48.8 \\
 \gr
 CAFormer-S36 & 811 & 7.1 & \textbf{53.2} & 72.1 & 57.7 & \textbf{46.0} & 69.5 & 49.8 \\
 \hline
 Swin-B               & 982 & 7.7   & 51.9 & 70.5 & 56.4 & 45.0 & 68.1 & 48.9 \\
 ConvNeXt-B             & 964 & 8.2   & 52.7 & 71.3 & 57.2 & 45.6 & 68.9 & 49.5 \\
 \gr
 ConvFormer-M36 & 912 & 6.7 & 53.0 & 71.4 & 57.4 & 45.7 & 69.2 & 49.5 \\
 \gr
 CAFormer-M36 & 920 & 6.4 & \textbf{53.8} & 72.5 & 58.3 & \textbf{46.5} & 70.1 & 50.7 \\
 \whline
\end{tabular}
\end{table}

\begin{table}[t]
\caption{\textbf{Performance of Semantic segmentation with UperNet \cite{upernet} on ADE20K~\cite{ade20k} validation set.} Images are cropped to $512 \times 512$ for training. The MACs are measured with input size of $512 \times 2048$. The FPS are measured on NVIDIA V100 GPU.}
\label{tab:upernet}
\centering
\begin{tabular}{l|cccc}
\whline
\multirow{2}{*}{Backbone} & \multicolumn{3}{c}{UperNet}\\
\cline{2-5}
&  Params (M) & MACs (G) & FPS & mIoU (\%) \\
    \whline
    Swin-T~\cite{swin}      & 60 & 945 & 21.3 & 45.8 \\
    ConvNeXt-T ~\cite{convnext}    & 60 & 939 & 21.3 & 46.7 \\
    \gr 
    ConvFormer-S18 & 54 & 925 & 23.7 & 48.6 \\
    \gr
    CAFormer-S18 & 54 & 1024 & 21.4 & \textbf{48.9} \\
	\hline
    Swin-S~\cite{swin}             & 81 & 1038 & 14.7 & 49.5 \\
    ConvNeXt-S~\cite{convnext}        & 82 & 1027 & 15.7 & 49.6 \\
    \gr
    ConvFormer-S36 &  67 & 1003 & 11.9 & 50.7 \\
    \gr
    CAFormer-S36 &  67 & 1197 & 10.8 & \textbf{50.8} \\
	\hline
    Swin-B~\cite{swin}      & 121 & 1188 & 14.6 & 49.7 \\
     ConvNeXt-B ~\cite{convnext}    & 122 & 1170 & 15.0 & 49.9 \\
     \gr
     ConvFormer-M36 & 85 & 1113 & 11.5 & 51.3 \\
     \gr
     CAFormer-M36  & 84 & 1346 & 9.8 & \textbf{51.7} \\
     
\whline
\end{tabular}

\normalsize
\end{table}

\subsection{Benchmark speed}

We first benchmark the speed of the proposed StarReLU and the commonly-used GELU \cite{gelu} on NVIDIA A100 GPU that is shown in Table \ref{tab:speed_act}. It can be seen that compared with GELU (Equation \ref{eqn:gelu}), StarReLU enjoys significant speedup, with $1.7\times$ speedup on NVIDIA A100 GPU. We also note that StarReLU is slower than GELU (PyTorch API) because the current implementation of StarReLU is not CUDA-optimized. 
Once optimized, we expect a further speedup for StarReLU, likely a significant one.

\begin{figure}[h]
  \centering
   \includegraphics[width=1\linewidth]{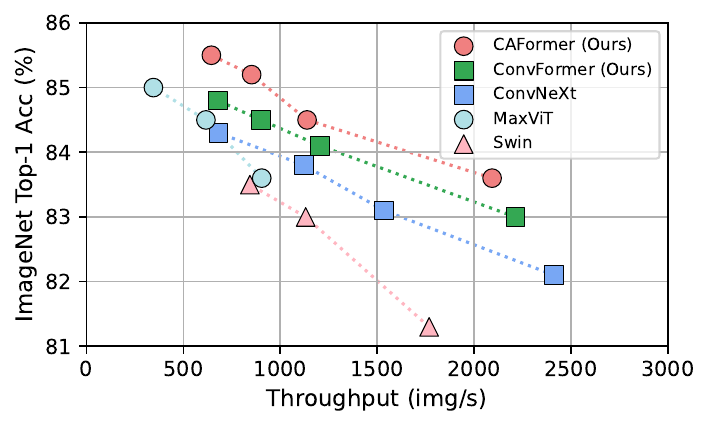}
   \caption{\textbf{Trade-off between accuracy and inference throughput.} The throughputs are measured on an NVIDIA A100 GPU with batch size of 128 and TF32.} 
   \label{fig:throughput}
\end{figure}

Then we further benchmark the ConvFormer, CAFormer, and other strong models (Swin \cite{swin} and ConvNeXt \cite{convnext}). For fair comparison, we replace StarReLU in ConvFormer and CAFormer with GELU. The results are shown in Table \ref{tab:speed_model}. We can see that for similar model sizes and MACs, ConvNeXt has the highest throughput. This is because ConvNeXt block has only one residual connection while MetaFormer block has two. However, ConvFormer and CAFormer obtain higher accuracy among these models and also achieve relatively higher throughputs than Swin and MaxViT, resulting in better trade-off between accuracy and throughput, as shown in Figure \ref{fig:throughput}.

\begin{table}[!t]
\caption{
\textbf{Benchmarking speed of activations.} We benchmark the speed by 10K runs with input shape of $1\times 1M$ in PyTorch \cite{paszke2019pytorch} on an NVIDIA A100 GPU. Note that it is unfair to directly compare GELU (PyTorch API) and StarReLU because the former is CUDA optimized while the last is not. 
}
\label{tab:speed_act}
\centering
\begin{tabular}{l|c}
\whline
Activation & Speed (runs/s) \\
\whline
GELU (PyTorch API) & \textcolor{gray}{145,067 (CUDA optimized)}  \\
GELU (Equation \ref{eqn:gelu}) & 20,273 \\
StarReLU & 35,257 (1.7x) \\
\whline
\end{tabular}

\end{table}

\begin{table}[!t]
\addtolength{\tabcolsep}{-5pt}
\caption{
\textbf{Inference throughputs of different models.} Models are trained and tested on resolution of $224^2$. We benchmark the throughputs on an NVIDIA A100 GPU with batch size of 128 and TF32.} 
\label{tab:speed_model}
\centering
\begin{tabular}{l | c c c c}
\whline
Models & Params (M) & MACs (G) &  Top-1 (\%)  & Throughput (img/s) \\
\whline
Swin-T & 29 & 4.5 & 81.3 & 1768  \\
ConvNeXt-T & 29 & 4.5 & 82.1  & 2413 \\
\gr
ConvFormer-S18 & 27 & 3.9 & 83.0  & 2213 \\
\gr
CAFormer-S18 & 26 & 4.1 & 83.6  & 2093 \\
Swin-S & 50 & 8.7 & 83.0 & 1131 \\
ConvNeXt-S & 50 & 8.7 & 83.1 & 1535 \\
MaxViT-T & 31 & 5.6 &  83.6 & 904 \\
\gr
ConvFormer-S36 & 40 & 7.6 & 84.1 & 1205 \\
\gr
CAFormer-S36 & 39 & 8.0 & 84.5  & 1138 \\
Swin-B & 88 & 15.4 & 83.5 & 843 \\
ConvNeXt-B & 89 & 15.4 & 83.8 & 1122 \\
MaxViT-S & 69 & 11.7 & 84.5 & 616 \\
\gr
ConvFormer-M36 & 57 & 12.8 & 84.5 & 899 \\
\gr
CAFormer-M36 & 56 & 13.2 & 85.2 & 852 \\
ConvNeXt-L & 198 & 34.4 & 84.3 & 681 \\
MaxViT-B & 120 & 23.4 & 85.0 & 345 \\
\gr
ConvFormer-B36 & 100 & 22.6 & 84.8 & 677 \\
\gr
CAFormer-B36 & 99 & 23.2 & 85.5 & 644 \\
\whline 
\end{tabular}

\end{table}

\section{Related work}
Transformer, since being introduced in~\cite{transformer}, has become a popular backbone for various tasks in NLP \cite{gpt1, gpt3, liu2019roberta, yan2019tener, raffel2020exploring}, computer vision \cite{igpt, vit, deit, t2t, pvt, swin} and other domains \cite{gulati2020conformer, pham2019very, li2019neural, kim2020t, huang2018music, li2019visualbert, chen2020uniter}. In computer vision, iGPT \cite{igpt} and ViT \cite{vit} introduce pure Transformer for self-supervised learning and supervised learning, attracting great attention in the research community to further improve Transformers. The success of Transformers had long been attributed to the attention module, and thus many research endeavors have been focused on improving attention-based token mixers \cite{t2t, pvt, swin}. However, it is shown in MLP-Mixer \cite{mlp-mixer} and FNet \cite{lee2021fnet} that, by replacing attention in Transformer with spatial MLP \cite{tay2021synthesizer} and Fourier transform, the resulting models still deliver competitive results. Along this line, \cite{metaformer} abstracts the Transformer into a general architecture termed MetaFormer, and meanwhile proposes the hypothesis that, it is the  MetaFormer that really plays a critical role in achieving promising performance. To this end, \cite{metaformer} specifies the token mixer to be as embarrassingly simple as pooling, and observes that the resultant model PoolFormer surpasses the well-tuned ResNet/ViT/MLP-like baselines \cite{resnet, rsb, vit, deit, pvt, mlp-mixer, liu2021pay, resmlp}. The power of MetaFormer can also be verified by the recent models adopting MetaFormer as the general architecture but with different attention-based \cite{li2022mvitv2, ren2022shunted, yang2021focal}, MLP-based \cite{mlp-mixer, resmlp, hou2022vision, chen2021cyclemlp, lian2021mlp}, convolution-based \cite{ding2022scaling, guo2022visual, yang2022focal, rao2022hornet, wu2019pay}, hybrid \cite{si2022inception, dai2021coatnet, li2022uniformer, tu2022maxvit} or other types of \cite{tatsunami2022sequencer, han2022vision} token mixers. Unlike these works, we do not attempt to introduce novel token mixers, but merely specify token mixers as the most basic or commonly-used operators to probe the capacity of MetaFormer.

\section{Conclusion}
In this paper, we make our exploration to study the capacity of MetaFormer, the abstracted architecture of Transformer. We take our eyes off the token-mixer design, and merely rely on the most basic or ``old-fashioned'' token mixers dated back years ago to build the MetaFormer models, namely IdentityFormer, RandFormer, ConvFormer, and CAFormer. The former two models, built upon identify mapping and randomized mixing, demonstrate the solid lower bound of MetaFormer and its universality to token mixers; the latter two models, built upon conventional separable convolutions and vanilla self-attention, readily offer recording-setting results. In our investigation, we also discover that a new activation, StarReLU, not only achieves better performance but also greatly reduces FLOPs of the activation function when compared with GELU. We expect MetaFormer to find its even broader domain of vision applications in future work, and cheerfully invite readers to try out the proposed MetaFormer baselines.

\ifCLASSOPTIONcompsoc
  \section*{Acknowledgments}
\else
  \section*{Acknowledgment}
\fi

This project is supported by
the Advanced Research and Technology Innovation Centre (ARTIC), the National University of Singapore (project number: A-0005947-21-00, project reference: ECT-RP2),
the Singapore Ministry of Education Academic Research Fund Tier~1 (WBS: A-0009440-01-00),
and the National Research Foundation Singapore under its AI Singapore Programme (Award Number: AISG2-RP-2021-023).
Weihao Yu and Xinchao Wang would like to thank TRC program and GCP research credits for the support of partial computational resources. We would like to thank Fredo Guan (independent researcher) and Ross Wightman (Hugging Face) for merging MetaFormer code into pytorch-image-models codebase.

\appendices
\section{Expectation and variance of Squared ReLU}
Assuming the input $x$ of Squared ReLU \cite{so2021primer} follows normal distribution with mean 0 and variance 1, \ie  $x \sim N(0, 1)$,  we have:
\begin{align}
\mathrm{E}(x^2) &= \mathrm{Var}(x) = 1 \\
\mathrm{E}\left((\mathrm{ReLU}(x))^2\right) &= \frac{1}{2} \mathrm{E}(x^2) = 0.5 \\
\mathrm{E}(x^4) &= \frac{1}{\sqrt{2 \pi}}\int_{-\infty}^{+\infty}z^4\exp\left(-\frac{z^2}{2}\right)dz \\
&= -\frac{1}{\sqrt{2\pi}}\int_{-\infty}^{+\infty}z^3d\left(\exp\left(-\frac{z^2}{2}\right)\right) \\
&=\left(-z^3\frac{1}{\sqrt{2\pi}}\exp\left(-\frac{z^2}{2}\right)\right)\bigg|_{-\infty}^{+\infty} +\\ &3\int_{-\infty}^{+\infty}z^2\frac{1}{\sqrt{2\pi}}\exp\left(-\frac{z^2}{2}\right)dz \\
&= 0 + 3\mathrm{E}(x^2) = 3 \\
\mathrm{E}\left((\mathrm{ReLU}(x))^4\right) &= \frac{1}{2}\mathrm{E}(x^4) = 1.5 \\
\mathrm{Var}\left((\mathrm{ReLU}(x))^2\right) &= \mathrm{E}\left((\mathrm{ReLU}(x))^4\right) - \left(\mathrm{E}\left((\mathrm{ReLU}(x))^2\right)\right)^2 \\
&= 1.5 - 0.5^2 = 1.25
\end{align}
where $\mathrm{E}(\cdot)$ and $\mathrm{Var}(\cdot)$ denote expectation and variance, respectively. Thus, we can obtain the expectation $\mathrm{E}\left((\mathrm{ReLU}(x))^2\right) = 0.5$ and variance $\mathrm{Var}\left((\mathrm{ReLU}(x))^2\right)= 1.25$.

\section{Code of separable convolution}
Algorithm \ref{alg:sep_conv} shows the PyTorch-like code of inverted separable convolution from MobileNetV2 \cite{mobilenetv2}.
\begin{algorithm}[!ht]
\caption{Token mixer of separable convolution, PyTorch-like Code}
\label{alg:sep_conv}
\definecolor{codeblue}{rgb}{0.25,0.5,0.5}
\definecolor{codekw}{rgb}{0.85, 0.18, 0.50}
\lstset{
  backgroundcolor=\color{white},
  basicstyle=\fontsize{7.5pt}{7.5pt}\ttfamily\selectfont,
  columns=fullflexible,
  breaklines=true,
  captionpos=b,
  commentstyle=\fontsize{7.5pt}{7.5pt}\color{codeblue},
  keywordstyle=\fontsize{7.5pt}{7.5pt}\color{codekw},
  showstringspaces=false,
  stringstyle=\color{darkred},
}
\begin{lstlisting}[language=python]
import torch
import torch.nn as nn
        
# Separable convolution
class SepConv(nn.Module):
    "Inverted separable convolution from MobileNetV2"
    def __init__(self, dim, kernel_size=7, padding=3, expansion_ratio=2, act1=nn.ReLU, act2=nn.Identity, 
        bias=False):
        super().__init__()
        med_channels = int(expansion_ratio * dim)
        self.pwconv1 = nn.Linear(dim, med_channels, bias=bias) # pointwise conv implemented by FC
        self.act1 = act1()
        self.dwconv = nn.Conv2d(med_channels, med_channels, kernel_size=kernel_size,
            padding=padding, groups=med_channels, bias=bias) # depthwise conv
        self.act2 = act2()
        self.pwconv2 = nn.Linear(med_channels, dim, bias=bias) # pointwise conv implemented by FC

    def forward(self, x):
        # [B, H, W, C] = x.shape
        x = self.pwconv1(x)
        x = self.act1(x)
        x = x.permute(0, 3, 1, 2) # [B, H, W, D] -> [B, D, H, W]
        x = self.dwconv(x)
        x = x.permute(0, 2, 3, 1) # [B, D, H, W] -> [B, H, W, D]
        x = self.act2(x)
        x = self.pwconv2(x)
        return x
\end{lstlisting}
\end{algorithm}

\vspace{-1mm}
\section{Hyper-parameters}
\vspace{-1mm}
The hyper-parameters of IdentityFormer, RandFormer and PoolFormerV2 trained on ImageNet-1K \cite{imagenet} are shown in Table \ref{tab:hyperparam_baisc_models}, while those of ConvFormer and CAFormer are shown in Table \ref{tab:hyperparam_convformer_caformer} for training on ImageNet-1K and Table \ref{tab:hyperparam_in21k} for pre-training on ImageNet-1K and fine-tuning on ImageNet-1K.

\begin{table*}[h]
\renewcommand{\arraystretch}{1.3}
  \caption{
  Hyper-parameters of IdentityFormer, RandFormer, PoolFormerV2 trained on ImageNet-1K.}
  \label{tab:hyperparam_baisc_models}
  \centering
  \begin{tabular}{l|ccc}
\whline
Hyper-parameter & IdentityFormer & RandFormer & PoolFormerV2 \\
\whline
Model size & \multicolumn{3}{c}{S12/S24/S36/M36/M48} \\
Epochs & \multicolumn{3}{c}{300} \\
Resolution & \multicolumn{3}{c}{$224^2$} \\
Batch size & \multicolumn{3}{c}{4096} \\
Optimizer & \multicolumn{3}{c}{AdamW} \\
Learning rate & \multicolumn{3}{c}{4e-3} \\
Learning rate decay & \multicolumn{3}{c}{Cosine} \\
Warmup epochs & \multicolumn{3}{c}{5} \\
Weight decay & \multicolumn{3}{c}{0.05} \\
Rand Augment & \multicolumn{3}{c}{9/0.5} \\
Mixup & \multicolumn{3}{c}{0.8} \\
Cutmix & \multicolumn{3}{c}{1.0} \\
Erasing prob & \multicolumn{3}{c}{0.25} \\
Peak stochastic depth rate & 0.1/0.1/0.2/0.3/0.4 & 0.1/0.1/0.2/0.3/0.3 & 0.1/0.1/0.2/0.3/0.4 \\
Label smoothing & \multicolumn{3}{c}{0.1} \\
\whline
\end{tabular}

\end{table*}

\begin{table*}[h]
\renewcommand{\arraystretch}{1.3}
  \caption{
  Hyper-parameters of ConvFormer and CAFormer trained on ImageNet-1K and finetuned at larger resolution of $384^2$.}
  \label{tab:hyperparam_convformer_caformer}
  \centering
  \begin{tabular}{l|cccc}
\whline
\multirow{2}{*}{Hyper-parameter} & \multicolumn{2}{c}{ConvFormer} & \multicolumn{2}{c}{CAFormer} \\
& Train & Finetune & Train & Finetune \\
\whline
Model size & \multicolumn{4}{c}{S18/S36/M36/B36} \\
Epochs & 300 & 30 & 300 & 30 \\
Resolution & $224^2$ & $384^2$ & $224^2$ & $384^2$ \\
Batch size & 4096 & 1024 & 4096 & 1024 \\
Optimizer & AdamW & AdamW & LAMB & LAMB \\
Learning rate & 4e-3 & 5e-5 & 8e-3 & 1e-4 \\
Learning rate decay & Cosine & None & Cosine & None \\
Warmup epochs & 20 & None & 20 & None \\
Weight decay & \multicolumn{4}{c}{0.05} \\
Rand Augment & \multicolumn{4}{c}{9/0.5} \\
Mixup & 0.8 & None & 0.8 & None \\
Cutmix & 1.0 & None & 1.0 & None \\
Erasing prob & \multicolumn{4}{c}{0.25} \\
Peak stochastic depth rate & 0.2/0.3/0.4/0.6 & 0.3/0.5/0.8/0.8 & 0.15/0.3/0.4/0.6 & 0.3/0.5/0.7/0.8 \\
MLP head dropout rate & 0/0/0/0 & 0.4/0.4/0.5/0.5 & 0/0.4/0.4/0.5 & 0.4/0.4/0.4/0.5 \\
Label smoothing & \multicolumn{4}{c}{0.1} \\
EMA decay rate & None & 0.9999 & None & 0.9999 \\
\whline
\end{tabular}

\end{table*}

\begin{table*}[h]
  \caption{
  Hyper-parameters of ConvFormer and CAFormer pre-trained on ImageNet-21K and fine-tuned on ImageNet-1K at resolution of $224^2$ and $384^2$.}
  \label{tab:hyperparam_in21k}
\renewcommand{\arraystretch}{1.3}
  \centering
  \begin{tabular}{l|cccc}
\whline
\multirow{2}{*}{Hyper-parameter} & \multicolumn{2}{c}{ConvFormer} & \multicolumn{2}{c}{CAFormer} \\
& Pretrain & Finetune & Pretrain & Finetune \\
\whline
Model size & \multicolumn{4}{c}{S18/S36/M36/B36} \\
Epochs & 90 & 30 & 90 & 30 \\
Resolution & $224^2$ & $224^2$/$384^2$ & $224^2$ & $224^2$/$384^2$ \\
Batch size & 4096 & 1024 & 4096 & 1024 \\
Optimizer & AdamW & AdamW & LAMB & LAMB \\
Learning rate & 1e-3 & 5e-5 & 2e-3 & 1e-4 \\
Learning rate decay & Cosine & None & Cosine & None \\
Warmup epochs & 5 & None & 5 & None \\
Weight decay & \multicolumn{4}{c}{0.05} \\
Rand Augment & \multicolumn{4}{c}{9/0.5} \\
Mixup & 0.8 & None & 0.8 & None \\
Cutmix & 1.0 & None & 1.0 & None \\
Erasing prob & \multicolumn{4}{c}{0.25} \\
Peak stochastic depth rate & 0/0/0.1/0.2 & 0.1/0.1/0.1/0.3 & 0.1/0.1/0.1/0.3 & 0.1/0.1/0.1/0.3 \\
MLP head dropout rate & 0.2/0.2/0.2/0.3 & 0.2/0.2/0.2/0.5 & 0.2/0.2/0.2/0.4 & 0.2/0.2/0.2/0.5 \\
Label smoothing & \multicolumn{4}{c}{0.1} \\
EMA decay rate & None & 0.9999 & None & 0.9999 \\
\whline
\end{tabular}

\end{table*}

\ifCLASSOPTIONcaptionsoff
  \newpage
\fi

\bibliographystyle{IEEEtran}
\bibliography{references}

\end{document}